\documentclass{llncs}
\usepackage{makeidx}  

\usepackage[utf8]{inputenc}

\usepackage[hidelinks]{hyperref}

\usepackage[numbers,sort&compress]{natbib}
\usepackage{url}
\usepackage{graphicx}
\usepackage[usenames,dvipsnames]{xcolor}
\usepackage[draft,inline,nomargin]{fixme}
\fxsetup{theme=color}

\usepackage{amsmath}
\usepackage{amssymb}
\usepackage{bbm}
\usepackage{textcomp}
\usepackage{siunitx}

\usepackage{booktabs}
\usepackage{threeparttable}
\usepackage{multirow}

\usepackage[capitalize]{cleveref}

\usepackage{tikz}
\usepackage{tikz-3dplot}
\usepackage[normalem]{ulem}   
\usetikzlibrary{arrows}
\usetikzlibrary{positioning,calc}
\usetikzlibrary{decorations.pathreplacing}
\usetikzlibrary{decorations.markings}
\usetikzlibrary{fit}
\usetikzlibrary{shapes.callouts}
\usetikzlibrary{shapes.geometric}
\usetikzlibrary{matrix}
\usetikzlibrary{spy}
\usepackage{setspace}

\usepackage{pgfplots}
\pgfplotsset{compat=1.9}
\usepgfplotslibrary{groupplots}
\usepgfplotslibrary{units}
\usepgfplotslibrary{colorbrewer}
\usepackage{colortbl}
\usepackage{pgfplotstable}

\usepackage{balance}

\usepackage{blindtext}

\usepackage{ifthen}
\usepackage{tabularx}
\usepackage{makecell}

\usepackage[export]{adjustbox}

\IfFileExists{graphicscache.sty}{\usepackage{graphicscache}}

\DeclareMathOperator*{\argmin}{arg\,min}
\DeclareMathOperator*{\concat}{concat} 

\usepackage{placeins}
\usepackage{dblfloatfix}
\newcommand{\rotateRPY}[3]
{   \pgfmathsetmacro{\rollangle}{#1}
    \pgfmathsetmacro{\pitchangle}{#2}
    \pgfmathsetmacro{\yawangle}{#3}

    \pgfmathsetmacro{\newxx}{cos(\yawangle)*cos(\pitchangle)}
    \pgfmathsetmacro{\newxy}{sin(\yawangle)*cos(\pitchangle)}
    \pgfmathsetmacro{\newxz}{-sin(\pitchangle)}
    \path (\newxx,\newxy,\newxz);
    \pgfgetlastxy{\nxx}{\nxy};

    \pgfmathsetmacro{\newyx}{cos(\yawangle)*sin(\pitchangle)*sin(\rollangle)-sin(\yawangle)*cos(\rollangle)}
    \pgfmathsetmacro{\newyy}{sin(\yawangle)*sin(\pitchangle)*sin(\rollangle)+ cos(\yawangle)*cos(\rollangle)}
    \pgfmathsetmacro{\newyz}{cos(\pitchangle)*sin(\rollangle)}
    \path (\newyx,\newyy,\newyz);
    \pgfgetlastxy{\nyx}{\nyy};

    \pgfmathsetmacro{\newzx}{cos(\yawangle)*sin(\pitchangle)*cos(\rollangle)+ sin(\yawangle)*sin(\rollangle)}
    \pgfmathsetmacro{\newzy}{sin(\yawangle)*sin(\pitchangle)*cos(\rollangle)-cos(\yawangle)*sin(\rollangle)}
    \pgfmathsetmacro{\newzz}{cos(\pitchangle)*cos(\rollangle)}
    \path (\newzx,\newzy,\newzz);
    \pgfgetlastxy{\nzx}{\nzy};
}

\tikzset{RPY/.style={x={(\nxx,\nxy)},y={(\nyx,\nyy)},z={(\nzx,\nzy)}}}

\newcommand{\Crossblue}{$\mathbin{\tikz [x=1.ex,y=1.ex,line width=.2ex, blue] \draw (0,0) -- (1,1) (0,1) -- (1,0);}$}%

\pgfplotstableset{
    /color cells/min/.initial=0,
    /color cells/max/.initial=1000,
    /color cells/textcolor/.initial=,
    color cells/.code={%
        \pgfqkeys{/color cells}{#1}%
        \pgfkeysalso{%
            postproc cell content/.code={%
                \begingroup
                \pgfkeysgetvalue{/pgfplots/table/@preprocessed cell content}\value
                \ifx\value\empty
                    \endgroup
                \else
                \pgfmathfloatparsenumber{\value}%
                \pgfmathfloattofixed{\pgfmathresult}%
                \let\value=\pgfmathresult
                \pgfplotscolormapaccess
                    [\pgfkeysvalueof{/color cells/min}:\pgfkeysvalueof{/color cells/max}]
                    {\value}
                    {\pgfkeysvalueof{/pgfplots/colormap name}}%
                \pgfkeysgetvalue{/pgfplots/table/@cell content}\typesetvalue
                \pgfkeysgetvalue{/color cells/textcolor}\textcolorvalue
                \toks0=\expandafter{\typesetvalue}%
                \xdef\temp{%
                    \noexpand\pgfkeysalso{%
                        @cell content={%
                            \noexpand\cellcolor[rgb]{\pgfmathresult}%
                            \noexpand\definecolor{mapped color}{rgb}{\pgfmathresult}%
                            \ifx\textcolorvalue\empty
                            \else
                                \noexpand\color{\textcolorvalue}%
                            \fi
                            \the\toks0 %
                        }%
                    }%
                }%
                \endgroup
                \temp
                \fi
            }%
        }%
    }
}

\begin{document}

\title{
YOLOPose V2: Understanding and Improving Transformer-based 6D Pose Estimation
}

\titlerunning{YOLOPose}  
%
\author{ Arul Selvam Periyasamy \and Arash Amini \and Vladimir Tsaturyan \and Sven Behnke}

\institute{Autonomous Intelligent Systems, University of Bonn, Germany\\
\email{periyasa@ais.uni-bonn.de}
}

\maketitle

\begin{abstract}
    6D object pose estimation is a crucial prerequisite for autonomous robot manipulation applications.
    The state-of-the-art models for pose estimation are convolutional neural network (CNN)-based.
    Lately, Transformers, an architecture originally proposed for natural language processing, is achieving state-of-the-art
    results in many computer vision tasks as well. Equipped with the multi-head self-attention mechanism, Transformers
    enable simple single-stage end-to-end architectures for learning object detection and 6D object pose estimation jointly.
    In this work, we propose YOLOPose (short form for You Only Look Once Pose estimation), 
    a Transformer-based multi-object 6D pose estimation method based on keypoint regression 
    and an improved variant of the YOLOPose model.
    In contrast to the standard heatmaps for predicting keypoints in an image, we directly regress the keypoints.
    Additionally, we employ a learnable orientation estimation module to predict the orientation from the keypoints.
    Along with a separate translation estimation module, our model is end-to-end differentiable. Our method
    is suitable for real-time applications and achieves results comparable to state-of-the-art methods. 
    We analyze the role of object queries in our architecture 
    and reveal that the object queries specialize in detecting objects in specific image regions.
    Furthermore, we quantify the accuracy trade-off of using datasets of smaller sizes to train our model.
\end{abstract}

Autonomous robotic object manipulation in real-world scenarios depends on high-quality 6D object pose estimation.
Such object poses are also crucial in many other applications like augmented reality, autonomous navigation, and industrial bin picking.
In recent years, with the advent of convolutional neural networks (CNNs), significant progress has been made 
to boost the performance of object pose estimation methods. Due to the complex nature of the task, 
the standard methods favor multi-stage approaches,
i.e., feature extraction followed by object detection and/or instance segmentation, target object crop extraction, and, finally,
6D object pose estimation. In contrast,
~\citet{carion2020end} introduced DETR, a Transformer-based single-stage architecture for object detection.
In our previous work~\citep{arash2021gcpr}, we extended the DETR model with the T6D-Direct architecture to perform multi-object 6D pose direct regression.
Taking advantage of the \textit{pleasingly parallel} nature of the Transformer architecture, the T6D-Direct model
predicts 6D pose for all the objects in an image in one forward-pass. Despite the advantages of the architecture and its
impressive performance, the overall 6D pose estimation accuracy of T6D-Direct, which directly regresses translation 
and orientation components of the 6D object poses, is inferior to state-of-the-art CNN-based methods, especially in rotation estimation.
Instead of directly regressing the translation and orientation components, the keypoint-based methods predict the 2D pixel
projection of 3D keypoints and use the perspective-\textit{n}-point (P\textit{n}P) algorithm to recover the 6D pose. In this work, we extend our T6D-Direct approach to utilize keypoints as 2D projected sparse correspondences. 
Our proposed model performs keypoint direct regression instead of the standard heatmaps for predicting the spatial position of the keypoints
in a given RGB image and uses a multi-layer perceptron (MLP) to learn the orientation component of 6D object pose from the keypoints. Another independent MLP serves as the translation direct estimator.
\noindent
In short, our contributions include:
\begin{enumerate}
 \item a Transformer-based real-time single-stage model for multi-object monocular 6D pose estimation using keypoint regression,
 \item a learnable rotation estimation module to estimate object orientation from a set of keypoints to develop an end-to-end differentiable architecture for pose estimation, and
 \item achieving results comparable to the state-of-the-art pose estimators on the YCB-Video dataset as well as yielding the fastest inference time.
\end{enumerate}

This article extends our conference paper~\citep{amini2022yolopose} that received the Best Paper Award at the 17th International Conference on Intelligent Autonomous Systems, 2021.
we make the following additional contributions:

\begin{enumerate}
 \item analyzing the role of object queries in the YOLOPose architecture, 
 \item improving the accuracy of the YOLOPose model by deriving new variants with additional inputs to the pose estimation MLPs,
 \item quantifying the robustness of the learned P\textit{n}P module compared to the analytical P\textit{n}P algorithm, and
 \item quantifying the accuracy trade-off of using datasets of smaller sizes to train our model.
\end{enumerate}

\begin{figure}[t]
        \centering
          \resizebox{.99\linewidth}{!}{\input{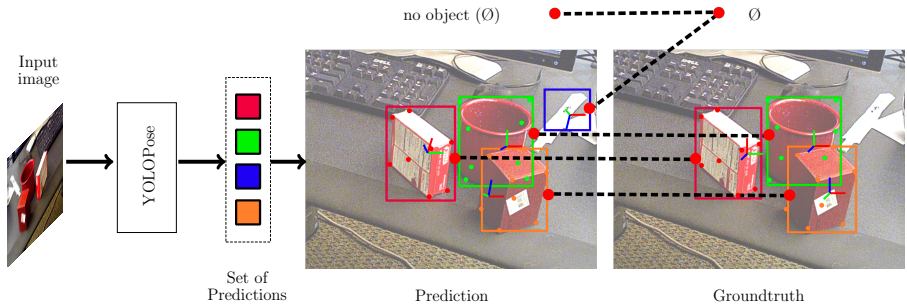}}
          \caption{Proposed YOLOPose approach. Our model predicts a set with a fixed cardinality. Each element in the set corresponds to an object prediction and after predicting all the objects in
          the given input image, the rest of the elements are padded with \text{\O} as no object predictions. The predicted and the ground-truth sets are matched using bipartite matching and the model is trained to minimize the Hungarian loss
          between the matched pairs. Our model is end-to-end differentiable.
          }
          \label{fig:pipeline}
\end{figure}

\section{Related Work}
\subsection{RGB Object Pose Estimation}
\label{sec:rgb_methods}

The recent significant progress in the task of 6D object pose estimation from RGB images is driven---like for many computer vision tasks---by deep learning methods. 
The current methods for object pose estimation from RGB images can be broadly classified into three major categories, namely direct regression methods, keypoint-based methods, and refinement-based methods.
Direct regression methods formulate the task of pose estimation as a regression of continuous translation and rotation components, 
whereas keypoint-based methods predict the location of projection of some of the specific keypoints or the 3D coordinates of the visible pixels of an object 
in an image and use the P\textit{n}P algorithm to retrieve the 6D  pose from the estimated 2D-3D correspondences. 
Often, the P\textit{n}P algorithm is used in conjunction with RANSAC for improving the robustness of the pose estimation.  

Some examples of direct regression methods include~\citep{xiang2017posecnn,periyasamy2018pose,Wang_2021_GDRN,arash2021gcpr}.
Keypoint-based include~\citep{rad2017bb8,tekin2018real,hu2019segmentation,peng2019pvnet,hu2020single}.
One important detail to note regarding these methods is that except for~\citep{arash2021gcpr,hu2019segmentation,thalhammer2021pyrapose,Capellen2019ConvPose} all the other methods use multi-staged CNNs. 
The first stage performs object detection and/or 
semantic or instance segmentation to detect the objects in the given RGB image. Using the object detections from the first stage, a 
crop containing the target object is extracted. In the second stage, these models predict the 6D pose of the target object. To enable end-to-end differentiability of the CNN models, these models 
employ region of interest (ROI) pooling, anchor box proposal, or non-maximum suppression (NMS) procedures~\citep{ren2015faster, hosang2017learning,redmon2017yolo9000}.
In terms of the 6D pose prediction accuracy, keypoint-based methods perform considerably better than the direct regression methods~\citep{hodavn2020bop}, 
though this performance gap is shrinking~\citep{arash2021gcpr}.

The third category of pose estimation methods are the refinement-based methods. These methods formulate the task of pose estimation as iterative pose refinement, 
i.e., the target object is rendered according to the current pose estimate, and a model is trained to estimate a pose update that minimizes the pose error between the ground-truth and the current pose prediction. Refinement-based methods~\citep{li2018deepim,manhardt2018deep,labbe2020,periyasamy2019refining} achieve the highest pose prediction accuracy among three categories~\citep{hodavn2020bop}. They need, however, a good object pose initialization within the basin of attraction of the final pose estimate.

\subsection{RGB-D Object Pose Estimation}
\label{sec:rgbd_methods}
Although we deal with the problem of RGB pose estimation in this work, it is highly relevant to review the RGB-D
methods as well. RGB-D deep learning methods for pose estimation fuse visual features from the RGB input extracted by a CNN model and geometric features from the point cloud or depth input. 
The predominant methods for extracting point-wise geometric features from the point cloud input include PointNet~\citep{NIPS2017_d8bf84be}, PointNet++~\citep{NIPS2017_d8bf84be}, and Point Transformer~\citep{zhao2021point}.
~\citet{xu2018pointfusion} learned to estimate 3D bounding box corners by fusing visual and geometric features.
~\citet{wang2019densefusion} learned dense point-wise embeddings from which the pose parameters are regressed in an iterative pose refinement procedure.
~\citet{he2020pvn3d} lifted the pixel-wise 2D keypoint offset learning proposed by~\citet{peng2019pvnet} for RGB images to 3D point clouds 
by learning point-wise 3D keypoint offset and using a deep Hough voting network.
~\citet{he2021ffb6d} jointly learned keypoint detection and instance segmentation and estimated 6D pose from the predicted
keypoint and segmentation using a least-squares fitting scheme from multi-view RGB-D input. 
Overall, RGB-D methods leverage the geometric features in the point cloud or depth input and achieve better accuracy than RGB-only methods.
Despite the advantages of the RGB-D data, the RGB-D sensors have limitations in terms of resolution and frame rate.
Reflectance and transparency properties of the objects also pose challenges for RGB-D sensors~\citep{kutulakos2008theory, li2020through, lysenkov2013recognition, Maeno_2013_CVPR}.
Additionally, calibrating RGB and depth sensors in large industrial settings is often time-consuming~\citep{basso2018robust, STARANOWICZ2015102, schwarz2018fast}.
Moreover, RGB methods are comparatively simpler and require less computational power and processing time. 
This motivates us in focusing on monocular RGB pose estimation.

\subsection{Learned P\textit{n}P}
Given a set of 3D keypoints and their corresponding 2D projections, and the camera intrinsics, the P\textit{n}P algorithm is used to recover the 6D object pose.
The standard P\textit{n}P algorithm~\citep{Gao2003CompleteSC} and its variant EP\textit{n}P~\citep{lepetit2009epnp} are used in combination with RANSAC to improve the robustness against outliers. Both P\textit{n}P and RANSAC are not trivially differentiable. In order to realize an end-to-end differentiable pipeline for the 6D object pose estimation,~\citet{Wang_2021_GDRN}, and~\citet{hu2020single} proposed a learning-based P\textit{n}P module. 
Similarly,~\citet{Li_2021_CVPR} introduced a learnable
3D Lifter module to estimate vehicle orientation. Recently,~\citet{chen2020end} proposed to differentiate  P\textit{n}P using the implicit function theorem. 
Although a generic differentiable P\textit{n}P 
has many potentials, due to the overhead incurred during training, we opt for a simple MLP that estimates the orientation component given the 2D keypoints.

\begin{figure*}
        \centering
          \resizebox{.99\linewidth}{!}{\input{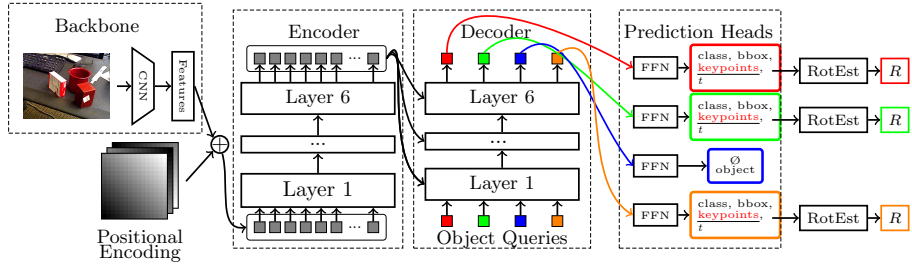}}
          \caption{YOLOPose architecture in detail. Given an RGB input image, we extract features using the standard ResNet model. 
          The extracted features are supplemented with positional encoding and provided as input to the Transformer encoder.
          The encoder module consists of six standard encoder layers with skip connections.
          The output of the encoder module is provided to the decoder module along with $N$ object queries. The decoder module
          also consists of six standard decoder layers with skip connections generating $N$ output embeddings. 
          The output embeddings are processed with FFNs to generate a set of $N$ elements in parallel. Each element in the set 
          is a tuple consisting of the bounding box, the class probability, the translation, and the interpolated bounding box keypoints.
          A learnable rotation estimation module is employed to estimate object orientation $R$ from the predicted 2D keypoints.
          }
          \label{fig:architecture}
\end{figure*}

\begin{figure}
    \centering
      \resizebox*{.75\linewidth}{4cm}{\input{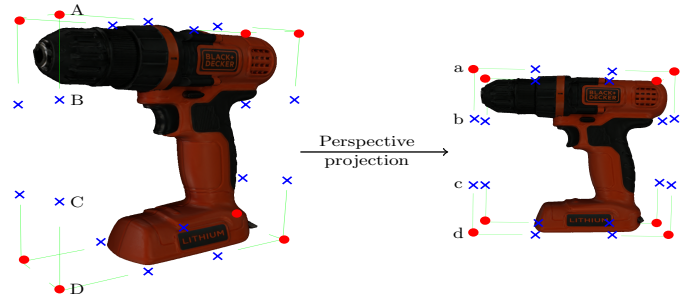}}
      \caption{Interpolated bounding box points. Bounding box points are indicated with red dots, and the interpolated points are indicated with blue crosses. The cross-ratio of every four collinear points is preserved during perspective projection, e.g., the cross-ratio of points A, B, C, and D remains the same in 3D and, after perspective projection, in 2D.
    }
      \label{fig:IBB}
\end{figure}

\section{Method}

\subsection{Multi-Object Keypoint Regression as Set Prediction}
\label{sec:set}
Object pose estimation is the task of estimating the position and the orientation of an object with respect to the sensor coordinate frame.
Occlusion, reflective properties of objects, lighting effects, and camera noise in real-world settings aggravate the complexity of the task.
The early methods for pose estimation like template matching~\citep{holzer2009distance, hinterstoisser2011gradient, cao2016real}
and keypoint-based~\citep{rothganger20063d, pavlakos20176, tulsiani2015viewpoints} decoupled object pose estimation from object detection and 
followed a multi-staged pipeline in which 2D bounding boxes are extracted in the first stage and only the crop containing the target object is processed 
in the second stage for pose estimation. Most of the deep learning methods also followed the same multi-stage approach for pose estimation. 
However, multi-stage pipelines suffer from two major issues. Firstly, inaccuracies in the first stage impede the final pose estimation accuracy, 
Secondly, complex modules like NMS, ROI, and anchor boxes are needed to realize end-to-end differentiable pipelines. 
Multi-object pose estimation methods alleviate the issues with multi-stage pipelines by detecting and localizing all objects in a given image. 
Following DETR~\citep{carion2020end} and T6D-Direct~\citep{arash2021gcpr}, we formulate multi-object pose estimation as a set prediction problem. 
\cref{fig:pipeline} gives an overview of our approach.
Given an RGB input image, our model outputs a set of elements with a fixed cardinality $N$. 
Each element in the set is a tuple containing the 2D bounding boxes, the class probability, 
the translation, and the keypoints. 2D bounding boxes are represented with the center coordinates, height, and width proportional to
the image size. The class probability is predicted using a softmax function.
To estimate translation $\mathbf{t}=[t_x, t_y, t_z]^T \in \mathbb{R}^3$ as the coordinate of the object origin in the camera coordinate system, we follow the method proposed by PoseCNN~\citep{xiang2017posecnn} 
which decouples the estimation of $\mathbf{t}$ into directly regressing the object's distance from the camera $t_z$ and the 2D location of projected 3D object's centroid in the image plane $[c_x, c_y]^T$. 
Finally, having the intrinsic camera matrix, we can recover $t_x$ and $t_y$. The exact choice of the keypoints is discussed in~\cref{sec:keypoints}.
The number of objects present in an image varies; therefore, to enable output sets with fixed cardinality, 
we choose $N$ to be larger than the expected maximum number of objects in an image in the dataset and introduce a no-object class \text{\O}. This \text{\O} class is analogous to the background class used in
semantic segmentation models. In addition to predicting the corresponding classes for objects present in the image, our model is trained to predict \text{\O} for the rest of the elements in the set.


\subsection{Model Architecture}
The proposed YOLOPose architecture is shown in~\cref{fig:architecture}.
The model consists of a ResNet backbone followed by Transformer-based encoder-decoder module and MLP prediction heads to predict a set of tuples described in~\cref{sec:set}.
CNN architectures have several inductive biases designed into them~\citep{lecun1995convolutional, CohenS17}.
These strong biases enable CNNs to learn efficient local spatial features in a fixed neighborhood defined by the receptive field
to perform well on many computer vision tasks. 
In contrast, Transformers, aided by the attention mechanism, are suitable for learning spatial features over the entire image. 
This makes the Transformer architecture suitable for multi-object pose estimation. In this section, we describe the individual components of the YOLOPose architecture.
\subsubsection{Backbone Network}
We use a ResNet50 backbone for extracting features from the given RGB image. For an image size of height H and width W, the backbone network extracts 2048 low-resolution feature maps of size 
H/32$\times$W/32. We then use 1$\times$1 convolution to reduce the 2048 feature dimensions to smaller \textit{d}=256 dimensions. The standard Transformer models are designed to process vectors. Hence, to enable processing the \textit{d}$\times$H/32$\times$W/32 features, we vectorize them to \textit{d}$\times\frac{H}{32}\frac{W}{32}$.

\subsubsection{Encoder}
The Transformer encoder module consists of six encoder layers with skip connections. Each layer performs multi-head self-attention of the input vectors. 
Given pixel with embedding $x$ of dimension $d$,
the embedding is split into $h$ chunks, or ``heads" and for each head $i$, the scaled dot-product attention is computed as:
$$\text{Attention}(Q,K,V) = \text{softmax}(\frac{QK^\top}{\sqrt{d/h}})V,$$
where $Q$, $K$, and $V$ are the query, key, and value matrices for the head $i$, respectively, and are computed by linearly projecting $x$ using projection parameter matrices $W^q$, $W^k$, $W^v$, respectively.
The attention outputs of the heads are concatenated and transformed linearly to compute MultiHead self-attention:
$$ \text{MultiHead(Q,K,V)}= \concat_{i \in h} (\text{Attention}(Q_i,K_i,V_i)) W^O,$$
where $W^O \in \mathbb{R}^{d \times d}$ is also a projection parameter matrix, and $\text{concat}$ denotes concatenation along the embedding dimension.
In contrast to the convolution operation, which limits the receptive field to a small neighborhood, self-attention
enables a receptive field of the size of the whole image. Note that the convolution operation can be cast as a special case of self-attention~\citet{Cordonnier2020On}.

\subsubsection{Positional Encodings}
The multi-head self-attention operation is permutation-invariant. 
Thus, the Transformer architecture ignores the order of the input vectors. 
We employ the standard solution of supplementing the input vectors with absolute positional encoding following ~\citet{carion2020end} to provide the Transformer model with spatial information of the pixels. We encode the pixel coordinates as sine and cosine functions of different frequencies:

$$ \text{P.E.}_{(pos, p)} = sin(pos/10000^{\frac{2p}{d}}), $$
$$ \text{P.E.}_{(pos, p+1)} = cos(pos/10000^{\frac{2p+1}{d}}), $$
where $pos$ is the pixel coordinate (either width or height), $d$ is the embedding dimension, and $p$ is the index of the positional encoding.
The positional embeddings are added to the backbone feature vectors before feeding them to the Transformer encoder as input. 

\subsubsection{Decoder}
On the decoder side, we compute cross-attention between the encoder output embeddings and $N$ learnable embeddings, referred to as \textit{object queries}, to generate decoder output embeddings, where $N$ is the cardinality of the predicted set.
The decoder consists of six decoder layers and the object queries are provided as input to each decoder layer.
Unlike the fixed  positional encoding used in the encoder, the object queries are learned jointly with the original learning objective---joint object detection and pose estimation, in our case---from the dataset. 
At the start of the training process, the object queries are initialized randomly, and during inference, the object queries are fixed.
In ~\cref{sec:obj_query_us}, we investigate the role of object queries generating object predictions.
The embeddings used in our model---both learned and fixed---are 256-dimensional vectors.

\subsubsection{FFN}
From the $N$ decoder output embeddings, we use feed-forward prediction heads
to generate a set of $N$ output tuples independently. 
Each tuple consists of the class probability, the bounding box, the keypoints, and the pose parameters. Prediction heads are fully-connected three-layer MLPs with hidden dimension 256 and ReLU activation in each layer.

\subsection{Keypoints Representation}
\label{sec:keypoints}
An obvious choice for selecting 3D keypoints is the eight corners of the 3D bounding box~\citep{oberweger2018}.
\citet{peng2019pvnet} instead used the Farthest Point Sampling (FPS) algorithm to sample eight keypoints on the surface of the object meshes, which are also spread out on the object to help the P\textit{n}P algorithm find a more stable solution.
~\citet{Li_2021_CVPR} defined the 3D representation of an object as sparse interpolated bounding boxes (IBBs), shown in~\cref{fig:IBB}, and exploited the property of perspective projection that cross-ratio of every four collinear points in 3D (A, B, C, and D as illustrated in~\cref{fig:IBB}) is preserved under perspective 
projection in 2D~\citep{hartley_zisserman_2004}. 
The cross-ratio consistency is enforced by an additional component in the loss function that the model learns to minimize during training. 
We further investigate these keypoints representations in~\cref{sec:ablation} and present our results in~\cref{tab:ablation}.

\subsection{RotEst}
\label{sec:rotest}
\begin{figure}
        \centering
          \begin{tikzpicture}[font=\sffamily\scriptsize]

\begin{scope}
\node [align=center, draw, fill=green!30, minimum width=52](in1) at (0, 0.8){IBB Keypoints};
\node [align=center, draw, fill=green!30, minimum width=52](in2) at (0, 0){Output\\Embedding };
\node [align=center, draw, fill=yellow!50](out) at (1.7, 0){FFN};
\draw[-{>[scale=0.3]}, line width=0.1pt] (in1.east) -- (out.west);
\draw[-{>[scale=0.3]}, line width=0.1pt] (in2.east) -- (out.west);

\node at (0, -1.5){(A)};
\end{scope}

\begin{scope}[xshift=4cm, yshift=0]
\node [align=center, draw, fill=green!30, minimum width=52](in1) at (0, 0.8){IBB Keypoints};
\node [align=center, draw, fill=green!30, minimum width=52](in2) at (0, 0){Output\\Embedding };
\node [align=center, draw, fill=green!30, minimum width=52](in3) at (0, -0.8){Canonical\\3D Points};
\node [align=center, draw, fill=yellow!50](out) at (1.7, 0){FFN};
\draw[-{>[scale=0.3]}, line width=0.1pt] (in1.east) -- (out.west);
\draw[-{>[scale=0.3]}, line width=0.1pt] (in2.east) -- (out.west);
\draw[-{>[scale=0.3]}, line width=0.1pt] (in3.east) -- (out.west);
\node at (0, -1.5){(B)};
\end{scope}

\begin{scope}[xshift=8cm, yshift=0]
\node [align=center, draw, fill=green!30, minimum width=52](in1) at (0, 0.8){IBB Keypoints};
\node [align=center, draw, fill=green!30, minimum width=52](in3) at (0, 0){Class\\Probabilities};
\node [align=center, draw, fill=yellow!50](out) at (1.7, 0){FFN};f
\draw[-{>[scale=0.3]}, line width=0.1pt] (in1.east) -- (out.west);
\draw[-{>[scale=0.3]}, line width=0.1pt] (in3.east) -- (out.west);
\node at (0, -1.5){(C)};
\end{scope}

\end{tikzpicture} 
          \caption{Variants of the YOLOPose model. All three variants are derived from the YOLOPose model and differ in the inputs provided to the pose estimation FFNs.
          }
          \label{fig:modifications}
\end{figure}
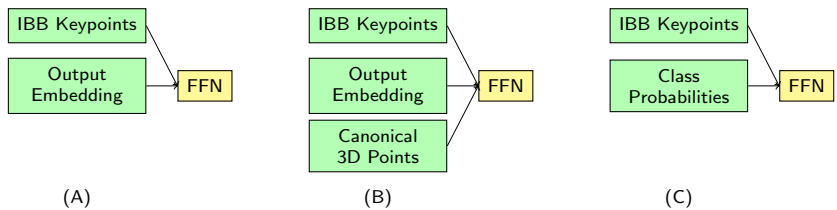

The standard solution for the perspective geometry problem of recovering 6D object/camera pose given 2D-3D correspondences and a calibrated camera is the P\textit{n}P algorithm.
 The minimum number of correspondences needed for employing  P\textit{n}P is 4. However, the accuracy and the robustness of the estimated pose increase with the number of correspondences. 
 Moreover, P\textit{n}P is used in conjecture with RANSAC to increase the robustness. 
Although P\textit{n}P is a standard and well-understood solution, incorporating it in neural network pipelines introduces two drawbacks.
First, it is not trivially differentiable. Second, P\textit{n}P combined with RANSAC needs multiple iterations to generate highly accurate pose predictions.
These drawbacks hinder us in realizing end-to-end differentiable pipelines with a single step forward pass for pose estimation.
To this end, we introduce the RotEst module.
For each object, from the estimated pixel coordinates onto which the 32 keypoints (the eight corners of the 3D bounding box and the 24 intermediate bounding box keypoints) are projected, the RotEst module predicts the object orientation represented as the 6D continuous representation in SO(3)~\citep{zhou2019continuity}. 
Furthermore, we experimented with providing additional inputs to the FFNs.
We created three variants of the YOLOPose model: variants A, B, and C (shown in \cref{fig:modifications}).
In variant A, in addition to the estimated IBB keypoints, we provide the output embedding of the object query to the FFNs.
In variant B, IBB keypoints, object query output embedding, and the canonical 3D bounding box points (based on the predicted object class) are provided to the FFNs,
whereas in variant C, estimated IBB keypoints and class probabilities are fed as input to FFNs.
Note that the size of the embedding used in YOLOPose model is 256. Thus, the number of parameters used in FFNs of the three variants is larger than that of the YOLOPose model.
We implement the RotEst module using six fully connected layers with a hidden dimension 1024 and a dropout probability of 0.5.

\subsection{Loss Function}
Our model is trained to minimize the Hungarian loss between the predicted and the ground-truth sets.
Computing the Hungarian loss involves finding the matching pairs in the two sets. 
We use bipartite matching~\citep{kuhn1955hungarian, stewart2016end, carion2020end} to find the permutation of the predicted elements
that minimize the matching cost.
Given the \text{\O} class padded ground-truth set $\mathcal{Y}$ of cardinality $N$ containing labels ${y}_1, {y}_2, ..., {y}_N$,
the predicted set denoted by $\hat{\mathcal{Y}}$, we search for the optimal permutation  
$\hat{\sigma}$ among the possible permutations $\sigma \in \mathfrak{S}_{N}$ 
that minimizes the matching cost $\mathcal{L}_{match}$.
Formally,
\begin{equation}
\hat{\sigma} = \argmin_{\sigma \in \mathfrak{S}_{N}} \sum_i^N  \mathcal{L}_{match}(y_i, \hat{y}_{\sigma(i)}).
\end{equation}

Although each element of the set is a tuple containing four components,
bounding box, class probability, translation, and keypoints, we use only the
bounding box and the class probability components to define the matching cost function. In
practice, omitting the other components in the cost function definition does not
hinder the model’s ability in learning to predict the keypoints and keeps the computational cost of the matching process minimal. 

Given the matching ground-truth and predicted sets $\mathcal{Y}$ and $\hat{\mathcal{Y}_\sigma}$, respectively, the Hungarian loss is computed as:
\begin{multline}\label{hungarian_loss}
\mathcal{L}_{Hungarian}(\mathcal{Y}, \hat{\mathcal{Y}_\sigma}) = \sum_i^N [-\text{log}\hat{p}_{\hat{\sigma}(i)}(c_i) +
                         \mathbbm{1}_{c_i\neq \text{\O}} \mathcal{L}_{box}(b_i, \hat{b}_{\hat{\sigma}(i)}) + \\
                         \lambda_{kp}\mathbbm{1}_{c_i\neq \text{\O}} \mathcal{L}_{kp}(k_i, \hat{k}_{\hat{\sigma}(i)}) +  
                         \lambda_{pose}\mathbbm{1}_{c_i\neq \text{\O}} \mathcal{L}_{pose}(R_i, t_i, \hat{R}_{\hat{\sigma}(i)}, \hat{t}_{\hat{\sigma}(i)})].
\end{multline}
\subsubsection{Class Probability Loss}
The class probability loss function is the standard negative log-likelihood. Since we choose the cardinality of the set to be higher than the expected maximum number of
objects in an image, the \text{\O} class appears disproportionately often. Thus, we weigh the loss for the \text{\O} class with a factor of 0.1.

\subsubsection{Bounding Box Loss}
The 2D bounding boxes are represented as $(c_x, c_y, w, h)$ where $(c_x, c_y)$ are 2D pixel coordinates and $w$ and $h$ are object width and height, respectively.
To train the bounding box prediction head,
We use a weighted combination of the Generalized IoU (GIoU)~\citep{rezatofighi2019generalized} and $\ell_1$-loss with 2 and 10 factors, respectively.

 \begin{equation}\label{eqn:box_loss}
 \mathcal{L}_{box}(b_i, \hat{b}_{\sigma(i)}) = \alpha \mathcal{L}_{iou}(b_i, \hat{b}_{\sigma(i)}) + \beta || b_i - \hat{b}_{\sigma(i)} ||,
 \end{equation}
 \begin{equation}
 \mathcal{L}_{iou}(b_i, \hat{b}_{\sigma(i)}) = 1 - \left( \frac{|b_i \cap \hat{b}_{\sigma(i)}|}{|b_i \cup \hat{b}_{\sigma(i)}|} - \frac{|B(b_i, \hat{b}_{\sigma(i)}) \setminus b_i \cup \hat{b}_{\sigma(i)} |}{|B(b_i, \hat{b}_{\sigma(i)})|}  \right),
\end{equation} and  $B(b_i, \hat{b}_{\sigma(i)})$ is the largest box containing both the ground truth $b_i$ and the prediction $\hat{b}_{\sigma(i)}$.

\subsubsection{Keypoint Loss}
\label{sec:kploss}

Having the ground truth $K_i$ and the model output $\hat{K}_{\hat{\sigma}(i)}$, the keypoints loss can be represented as: 
\begin{equation}\label{eqn:keypoints_loss}
    \mathcal{L}_{kp}(K_i, \hat{K}_{\hat{\sigma}(i)}) = \gamma ||K_i - \hat{K}_{\hat{\sigma}(i)}||_1 + \delta \mathcal{L_{CR}},
\end{equation}
where $\gamma$ and $\delta$ are hyperparameters. The first part of the keypoints loss is the $\ell_1$ loss, and for the second part, we employ the cross-ratio loss $\mathcal{L_{CR}}$ defined in Equation~\ref{equ:cr_loss} to enforce the cross-ratio consistency in the keypoint loss as proposed by~\citet{Li_2021_CVPR}. This loss is self-supervised by preserving the cross-ratio of each line to be 4/3. The reason is that after the camera projection of the 3D bounding box on the image plane, the cross-ratio of every four collinear points remains the same.

\begin{equation}
\label{equ:cr_loss}
    \mathcal{L_{CR}} = Smooth\ell_1(\text{CR}^2 -  \frac{||c-a||^2||d-b||^2}{||c-b||^2||d-a||^2}),~\text{CR} = \frac{||C-A||~||D-B||}{||C-B||~||D-A||} = \frac{4}{3},
\end{equation}
where $\text{CR}^2$ is chosen since $||.||^2$ can be easily computed using vector inner product. A, B, C, and D are four collinear points and their corresponding predicted 2D projections are a, b, c, and d, respectively.





\subsubsection{Pose Loss}
\label{sec:pose_loss}
We supervise the rotation $R$ and the translation $\mathbf{t}$ individually via employing PLoss and SLoss from~\citep{xiang2017posecnn} for rotation, and $\ell_1$ loss for translation.

\begin{equation}\label{eqn:ploss}
\mathcal{L}_{pose}(R_i, \mathbf{t}_i, \hat{R}_{\sigma(i)}, \hat{\mathbf{t}}_{\sigma(i)}) = \mathcal{L}_{rot}(R_i, \hat{R}_{\sigma(i)}) + || \mathbf{t}_i - \hat{\mathbf{t}}_{\sigma(i)} ||_1,
\end{equation}

\begin{equation}\label{eqn:pose_loss}
\mathcal{L}_{rot} = \left\{\begin{array}{ll}
\frac{1}{|\mathcal{M}_i|} \displaystyle\sum_{\text{x}_1 \in \mathcal{M}_i}  \min_{\text{x}_2 \in \mathcal{M}_i}|| R_i\text{x}_1 - \hat{R}_{\sigma(i)} \text{x}_2 ||_1 \text { if symmetric, } \\
\frac{1}{|\mathcal{M}_i|} \displaystyle\sum_{\text{x} \in \mathcal{M}_i} || R_i\text{x} - \hat{R}_{\sigma(i)} \text{x} ||_1 \text { otherwise, } 
\end{array}\right.
\end{equation}
where $\mathcal{M}_i$ indicates the set of 3D model points. Here, we subsample 1.5K points from meshes provided with the dataset. $R_i$ is the ground truth rotation, and $\mathbf{t}_i$ is the ground truth translation. $\hat{R}_{\sigma(i)}$ and $\hat{\mathbf{t}}_{\sigma(i)}$ are the predicted rotation and translation, respectively.

\section{Evaluation}
In this section, we evaluate the performance of our proposed YOLOPose model and its variants and compare it with other state-of-the-art 6D pose estimation methods.

\begin{figure*}
        \centering
        \newlength{\imgres}
        \setlength{\imgres}{0.16\textwidth}
        \setlength{\tabcolsep}{0.01cm}
        \begin{tabular}{cccccc}
            
         \includegraphics[width=\imgres]{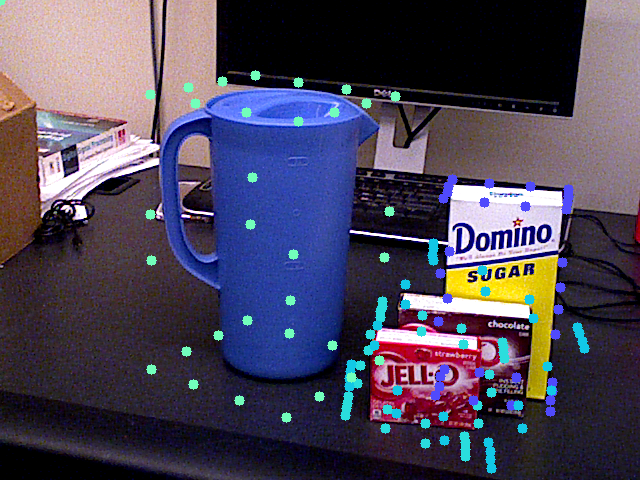} &
         \includegraphics[width=\imgres]{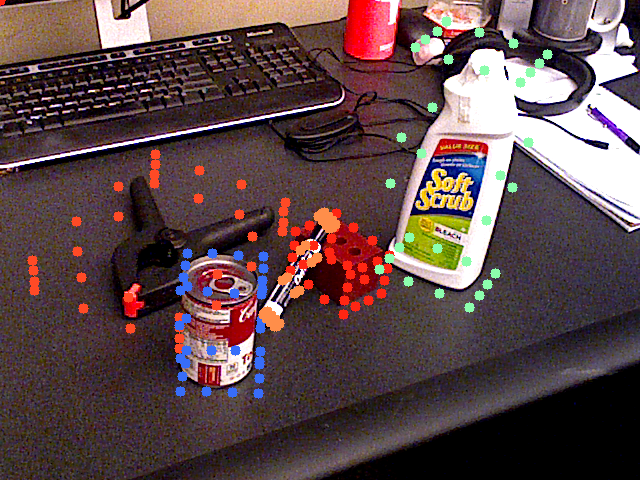} &
         \includegraphics[width=\imgres]{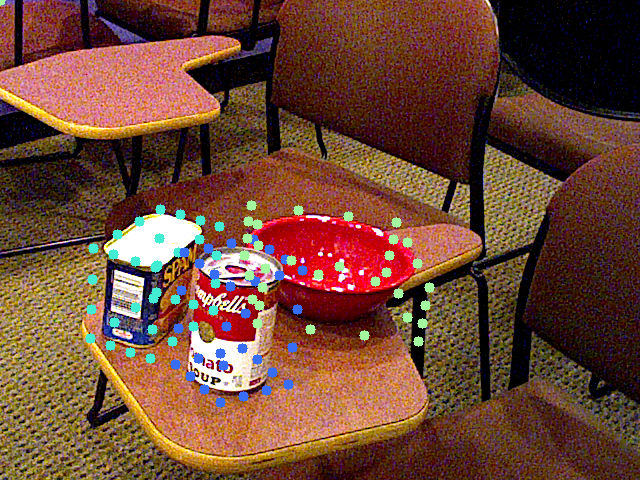} &
         \includegraphics[width=\imgres]{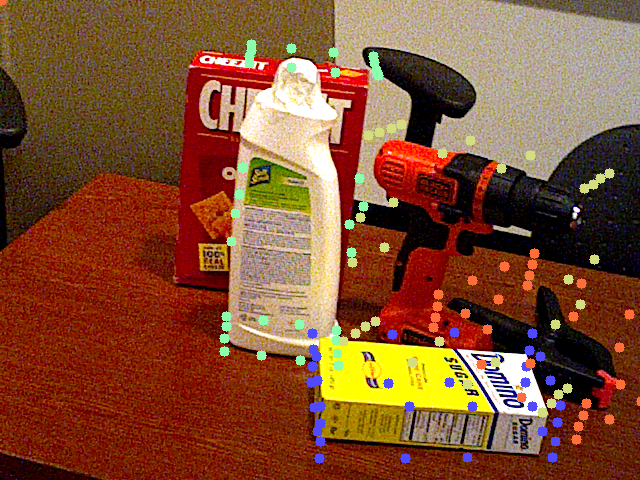} &
         \includegraphics[width=\imgres]{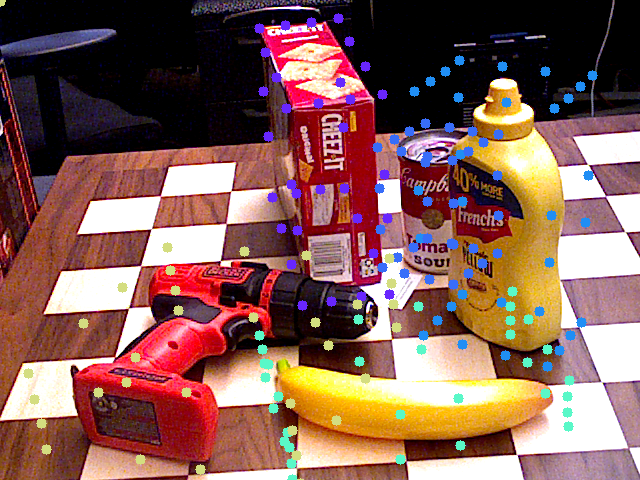} &
         \includegraphics[width=\imgres]{figures/yoloposeA/good_vis/78/key_000056_001046.png} \\
             
         \includegraphics[width=\imgres]{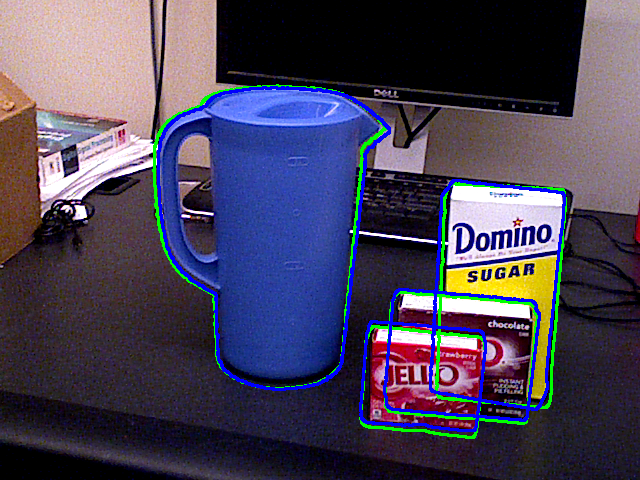} &
         \includegraphics[width=\imgres]{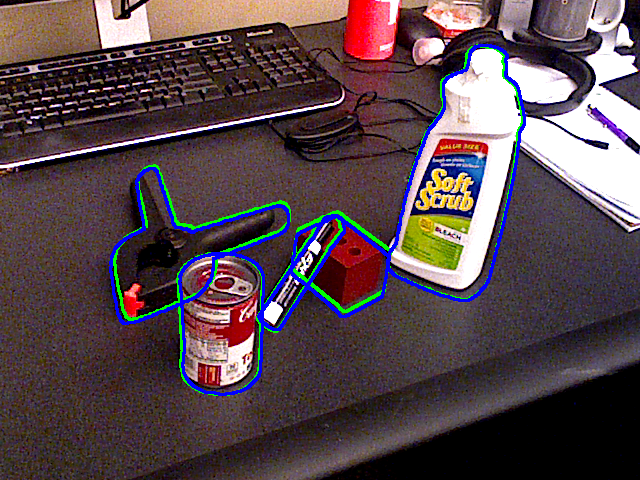} &
         \includegraphics[width=\imgres]{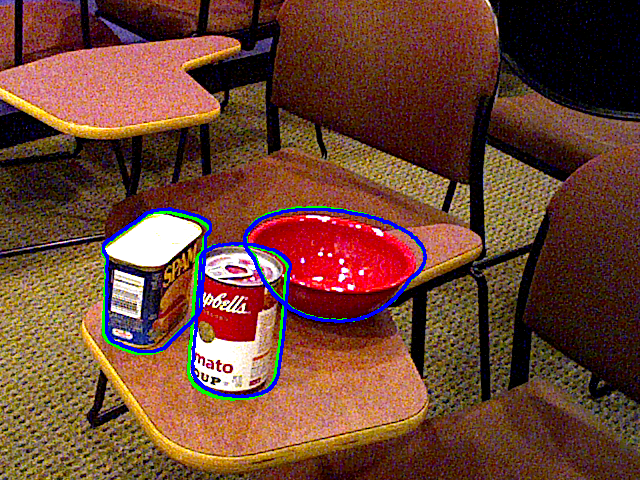} &
         \includegraphics[width=\imgres]{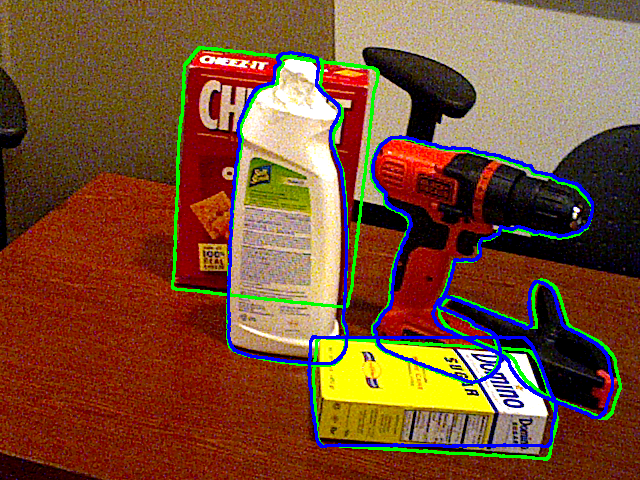} &
         \includegraphics[width=\imgres]{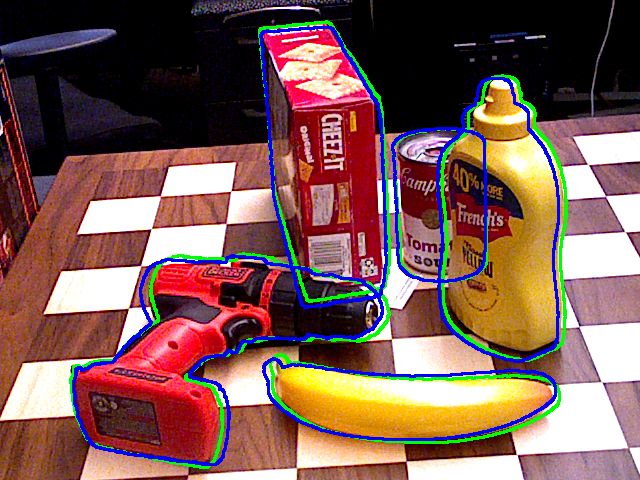} &
         \includegraphics[width=\imgres]{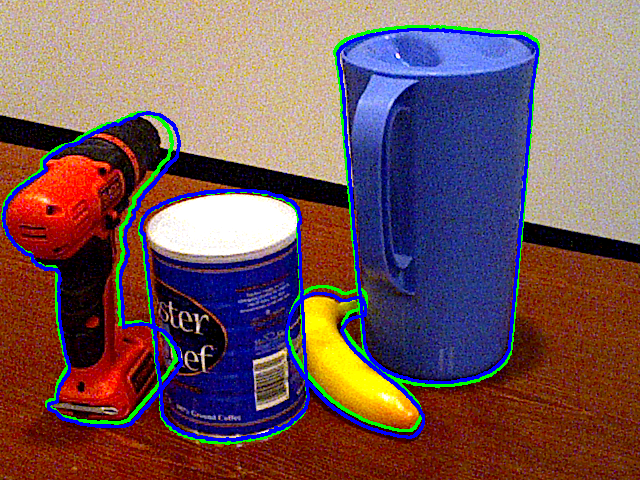} \\
        \end{tabular}
        \caption{Qualitative results on YCB-V test set. 
        Top row: The predicted IBB keypoints overlaid on the input images.
        Bottom row: Ground truth and predicted object poses are visualized as object contours in green and blue colors, respectively.}
        \label{fig:result}
\end{figure*}

\subsection{Dataset}
\label{sec:dataset}
We use the challenging YCB-Video (YCB-V)~\citep{xiang2017posecnn} dataset to evaluate the performance of our model.
YCB-V provides bounding box, segmentation, and 6D pose annotations for 133,936 RGB-D images.
Since our model is RGB-based, we do not use the provided depth information. 
The dataset is generated by capturing video sequences of a random subset of objects from a total of 21 objects placed in tabletop configuration.
From the 92 video sequences, twelve are used for testing and 80 are used for training.
The objects used exhibit varying geometric shapes, reflectance properties, and symmetry. Thus, YCB-V is a challenging dataset for benchmarking
6D object pose estimation methods.
YCB-V also provides high-quality meshes for all 21 objects.
Mesh points from these objects are used in computing the evaluation metrics that we describe in~\cref{sec:metric}.~\citet{hodavn2020bop} provided a variant of YCB-V\footnote{\url{https://bop.felk.cvut.cz/datasets/}} as  part of the BOP challenge in which the centers of the 3D bounding boxes are aligned with the origin of the model coordinate system and the ground-truth annotations are converted correspondingly. We use the BOP variant of the YCB-V dataset.
In addition to the YCB-V dataset images, we use the synthetic dataset provided by PoseCNN~\citep{xiang2017posecnn} for training our model.
Moreover, we initialize our model using the pre-trained weights on the COCO dataset~\citep{lin2014microsoft} for the task of object detection.

\subsection{Metrics}
\label{sec:metric}

\citet{xiang2017posecnn} proposed area under the curve (AUC) of ADD and ADD-S metrics for evaluating the accuracy of non-symmetric and symmetric objects, respectively.
Given the ground-truth 6D pose annotation with rotation and translation components $R$ and $\mathbf{t}$, and the predicted rotation and translation components 
$\hat{R}$ and $\hat{\mathbf{t}}$, ADD metric is the average $\ell_2$ distance between the subsampled mesh points $\mathcal{M}$ in the ground truth and the predicted pose. 
In contrast, the symmetry-aware ADD-S metric is the average distance between the closest subsampled mesh points $\mathcal{M}$ in the ground-truth and predicted pose.
Following the standard procedure proposed by~\citet{xiang2017posecnn}, we aggregate the results and report the area under the threshold-accuracy
curve for distance thresholds from zero to 0.1m.
\begin{equation}
        \text{ADD} = \frac{1}{|\mathcal{M}|} \sum_{x \in \mathcal{M}}\|(Rx+\mathbf{t})-(\hat{R} x+\hat{\mathbf{t}})\|,
\end{equation}

\begin{equation}
        \text{ADD-S} = \frac{1}{|\mathcal{M}|} \sum_{x_{1} \in \mathcal{M}} \min _{x_{2} \in \mathcal{M}}\|(R x_{1}+\mathbf{t})-(\hat{R} x_{2}+\hat{\mathbf{t}})\|.
\end{equation}

The ADD and ADD-S metrics are combined into one metric by using ADD for non-symmetric objects and ADD-S for symmetric objects. 
This combined metric is denoted as ADD-(S).

\subsection{Hyperparameters}
The $\gamma$ and $\delta$ hyperparameters in $\mathcal{L}_{kp}$ (\cref{eqn:keypoints_loss}) are set to 1 and 10, respectively.
While computing the Hungarian loss, the pose loss component is weighted down by a factor of 0.05. The cardinality of the predicted set $N$=20.
The model takes images of the size 640 $\times$ 480 as input and is trained using the AdamW optimizer~\cite{adamw}
with an initial learning rate of $10^{-4}$ for 150 epochs. Afterward, the model is trained additionally for 50 epochs, with a reduced learning rate by a factor of 0.1. The batch size is 32. Gradient clipping with a maximal gradient norm of 0.1 is applied.

\subsection{Results}

\begin{table}
\centering
\setlength{\aboverulesep}{0pt}
\setlength{\belowrulesep}{0pt}
\caption{Comparison of the proposed keypoints-based method YOLOPose with the state-of-the-art methods on YCB-V.  
The symmetric objects are denoted by *. The best results are shown in bold.}
\label{tab:ycbv-details-keypoints}
\resizebox{\textwidth}{!}{
\begin{tabular}{l|cc|cc|cc|cc|cc}
\toprule
Method &
\multicolumn{2}{c|}{\thead{GDR-Net \\~\citep{Wang_2021_GDRN}}} &
\multicolumn{2}{c|}{\thead{YOLOPose \\ (Ours)}} &
\multicolumn{2}{c|}{\thead{YOLOPose-A \\ (Ours)}} &
\multicolumn{2}{c|}{\thead{DeepIM \\~\citep{li2018deepim}}} &
\multicolumn{2}{c}{\thead{YOLOPose-A \\ (Ours)}}
 \\
\midrule
Metric & 
\thead{AUC of \\ ADD-S} &
\thead{AUC of \\ ADD(-S)} & 
\thead{AUC of \\ ADD-S} & 
\thead{AUC of \\ ADD(-S)} & 
\thead{AUC of \\ ADD-S} & 
\thead{AUC of \\ ADD(-S)} & 
\thead{AUC of \\ ADD-S} & 
\thead{AUC of \\ ADD(-S)} & 
\thead{AUC of \\ ADD-S \\@0.1d} & 
\thead{AUC of \\ ADD(-S) \\@0.1d}\\

\midrule
master\_chef\_can        & \textbf{96.6} & 71.1                   & 91.3          & 64.0             & 91.7            & \textbf{71.3} & 93.1          & 71.2            &   71.3        & 36.6 \\
cracker\_box             & 84.9          & 63.5                   & 86.8          & 77.9             & \textbf{92.0}   & 83.3          & 91.0          & \textbf{83.6}   &   83.3        & 71.1 \\
sugar\_box               & \textbf{98.3} & 93.2                   & 92.6          & 87.3             & 91.5            & 83.6          & 96.2          & \textbf{94.1}   &   83.6        & 59.5 \\
tomato\_soup\_can        & \textbf{96.1} & \textbf{88.9}          & 90.5          & 77.8             & 87.8            & 72.9          & 92.4          & 86.1            &   72.9        & 29.8 \\
mustard\_bottle          & \textbf{99.5} & \textbf{93.8}          & 93.6          & 87.9             & 96.7            & 93.4          & 95.1          & 91.5            &   93.4        & 93.4 \\
tuna\_fish\_can          & 95.1          & 85.1                   & 94.3          & 74.4             & 94.9            & 70.5          & \textbf{96.1} & \textbf{87.7}   &   70.5        & 17.4 \\
pudding\_box             & \textbf{94.8} & 86.5                   & 92.3          & 87.9             & 92.6            & \textbf{87.0} & 90.7          & 82.7            &   87.0        & 70.9 \\
gelatin\_box             & \textbf{95.3} & 88.5                   & 90.1          & 83.4             & 92.2            & 85.7          & 94.3          & \textbf{91.9}   &   85.7        & 23.4 \\
potted\_meat\_can        & 82.9          & 72.9                   & 85.8          & \textbf{76.7}    & 85.0            & 71.4          & \textbf{86.4} & 76.2            &   71.4        & 31.2 \\
banana                   & \textbf{96.0} & 85.2                   & 95.0          & 88.2             & 95.8            & \textbf{90.0} & 91.3          & 81.2            &   90.0        & 83.1 \\
pitcher\_base            & \textbf{98.8} & \textbf{94.3}          & 93.6          & 88.5             & 95.2            & 90.8          & 94.6          & 90.1            &   90.8        & 90.1 \\
bleach\_cleanser         & \textbf{94.4} & 80.5                   & 85.3          & 73.0             & 83.1            & 70.8          & 90.3          & \textbf{81.2}   &   70.8        & 62.9 \\
bowl$^*$                 & 84.0          & 84.0                   & 92.3          & 92.3             & \textbf{93.4}   & \textbf{93.4} & 81.4          & 81.4            &   93.4        & 87.4 \\
mug                      & \textbf{96.9} & 87.6                   & 84.9          & 69.6             & 95.5            & \textbf{90.0} & 91.3          & 81.4            &   90.0        & 71.0 \\
power\_drill             & 91.9          & 78.7                   & \textbf{92.6} & \textbf{86.1}    & 92.5            & 85.2          & 92.3          & 85.5            &   85.2        & 73.6 \\
wood\_block$^*$          & 77.3          & 77.3                   & 84.3          & 84.3             & \textbf{93.0}   & \textbf{93.0} & 81.9          & 81.9            &   93.0        & 93.0 \\
scissors                 & 68.4          & 43.7                   & \textbf{93.3} & \textbf{87.0}    & 80.9            & 71.2          & 75.4          & 60.9            &   71.2        & 42.5 \\
large\_marker            & \textbf{87.4} & 76.2                   & 84.9          & 76.6             & 85.2            & \textbf{77.0} & 86.2          & 75.6            &   77.0        & 14.7 \\
large\_clamp$^*$         & 69.3          & 69.3                   & 92.0          & 92.0             & \textbf{94.7}   & \textbf{94.7} & 74.3          & 74.3            &   94.7        & 94.1 \\
extra\_large\_clamp$^*$  & 73.6          & 73.6                   & \textbf{88.9} & \textbf{88.9}    & 80.7            & 80.7          & 73.3          & 73.3            &   80.7        & 65.7 \\
foam\_brick$^*$          & 90.4          & 90.4                   & 90.7          & 90.7             & \textbf{93.8}   & \textbf{93.8} & 81.9          & 81.9            &   93.8        & 78.9 \\
\midrule
MEAN                     & 89.1          & 80.2                   & 90.1          & 82.6             & \textbf{91.2}   & \textbf{83.3} & 88.1          & 81.9            &   83.3        & 61.4\\ 
\bottomrule
\end{tabular}
}
\label{tab:ycbv-details}
\end{table}

\begin{table}
      \centering
      \footnotesize
      \setlength{\aboverulesep}{0pt}
      \setlength{\belowrulesep}{0pt}
        \caption{Pose estimation results on YCB-V.
        }
\begin{tabular}{c|m{90pt}|c|c|c|c}
    \toprule
    Input & \thead{Method} & ADD(-S) & \thead{AUC of \\ ADD-S} & \thead{AUC of \\ ADD(-S)} & \thead{Inference \\Time \\{} [\si{\ms}/frame] }\\
    \midrule
    \multirow{5}{*}{RGB} & CosyPose$^{\dagger}$~\citep{labbe2020} & - & 89.8 & \textbf{84.5} & 395\\
    & PoseCNN~\citep{xiang2017posecnn} & 21.3 & 75.9 & 61.3 & -\\
    & GDR-Net~\citep{Wang_2021_GDRN} & 49.1 & 89.1 & 80.2 & 65\\
    & YOLOPose (Ours) &  65.0 & 90.1 & 82.6 & \textbf{17} \\
    & YOLOPose-A (Ours) &  \textbf{69.0} & \textbf{91.2} & 83.3 & 22 \\
    \midrule
    \multirow{4}{*}{RGB-D} & PVNet3D~\citep{he2020pvn3d} & - & 95.5 & 91.8 & 170 \\
    & PVNet3D+ICP~\citep{he2020pvn3d} & - & 96.1 & 92.3 & 190 \\
    & FFB6D~\citep{xiang2017posecnn} & - & 96.6 & 92.7 & \textbf{75} \\
    & FFB6D+ICP~\citep{xiang2017posecnn} & - & \textbf{97.0} & \textbf{93.7} & 95 \\
    \bottomrule 
  \end{tabular}
    \label{tab:inftime}
    \\
    $^{\dagger}$ indicates the refinement-based method.
\end{table}

\begin{figure}
        \centering
        \newlength{\imgw}
        \setlength{\imgw}{2.0cm}
        \setlength{\tabcolsep}{0.009cm}
        \resizebox{0.9\linewidth}{!}{
        \begin{tabular}{p{1.em}cccc}
                \raisebox{2.75\normalbaselineskip}[0pt][0pt] {\rotatebox[origin=c]{90}{\scriptsize Scene \qquad \qquad}} &
         \includegraphics[width=\imgw]{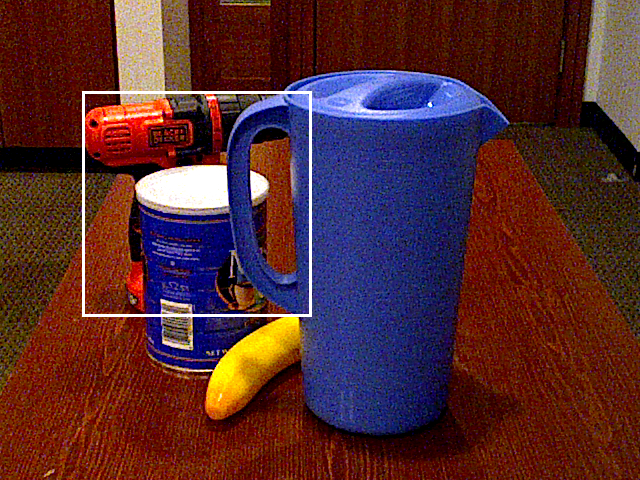} &
         \includegraphics[width=\imgw]{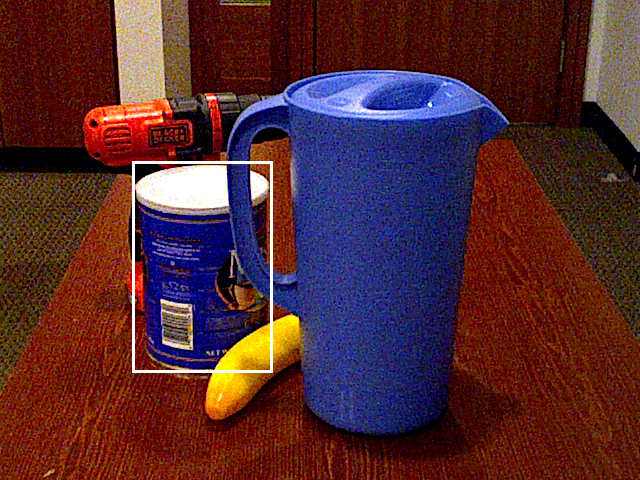} &
         \includegraphics[width=\imgw]{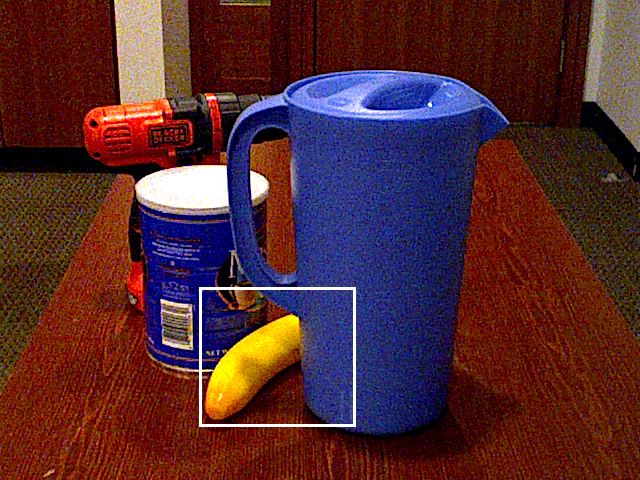} &
         \includegraphics[width=\imgw]{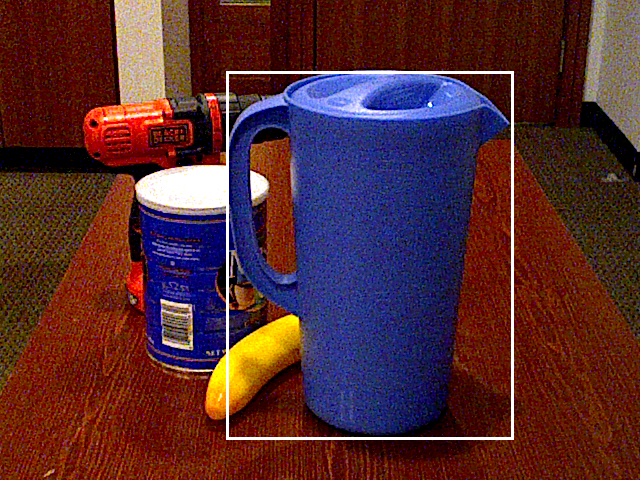} \\
         \raisebox{1.75\normalbaselineskip}[0pt][0pt] {\rotatebox[origin=c]{90}{\scriptsize Attention Maps}} &
         \includegraphics[width=\imgw]{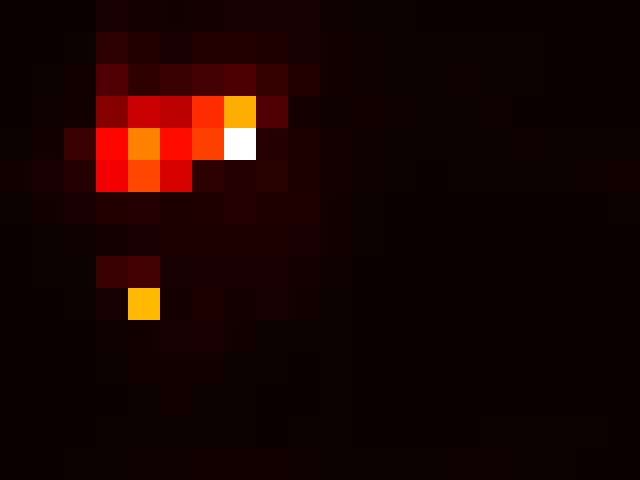} &
         \includegraphics[width=\imgw]{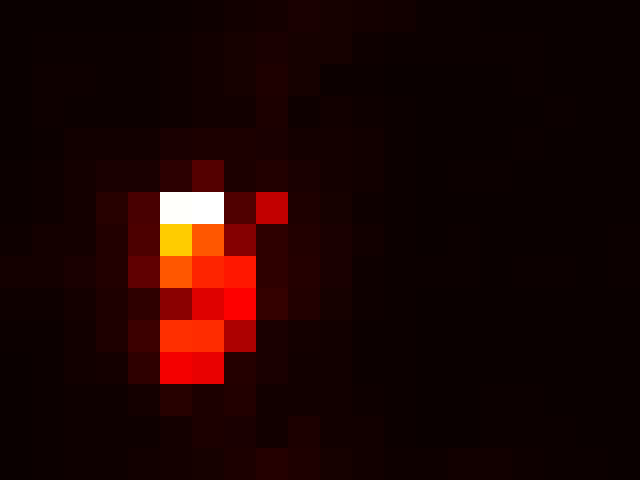} &
         \includegraphics[width=\imgw]{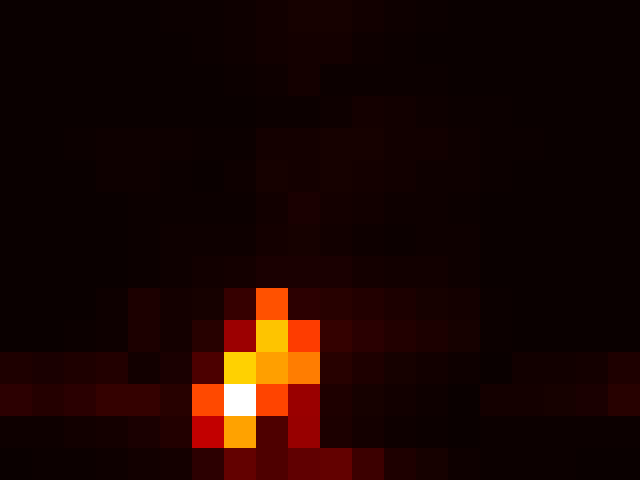} &
         \includegraphics[width=\imgw]{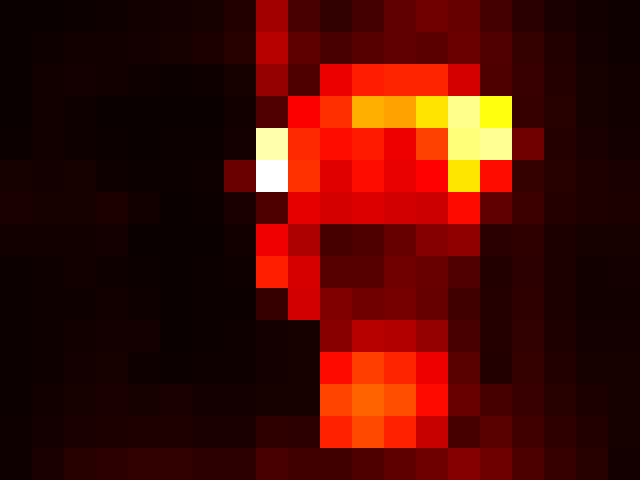} \\
         & (a) & (b) & (c) & (d) \\
        \end{tabular}
        }
        \tikz{
            \pgfplotscolorbardrawstandalone[ 
                colormap/hot2,
                point meta min=0,
                point meta max=1,
                colorbar horizontal, 
            colorbar style={
                height=0.15cm,
                width=3cm,
                font=\scriptsize
            }]
        }
        \caption{Top: Object detections predicted by bounding boxes in the given image.
        Bottom: Decoder cross-attention maps for the object queries corresponding to the predictions in the first row.
        }
        \label{fig:multiheadattn}
\end{figure}

In this section, we present the quantitative and qualitative results of the proposed method.
We present exemplar qualitative results in~\cref{fig:result}.
In~\cref{tab:ycbv-details}, we provide the quantitative per class area under the accuracy curve (AUC) of the ADD-S and ADD(-S) metrics. 
Both YOLOPose and YOLOpose-A perform well across all object categories and achieve higher AUC scores than the methods in comparison.
YOLOPose-A achieves an impressive AUC of ADD-S and ADD-(S) score of 91.2 and 83.3, respectively, which is an improvement of 1.1 and 0.7 over the YOLOPose model.
In terms of the individual objects, YOLOPose-A performs significantly better than the mean on \textit{mustard bottle},
\textit{bowl}, \textit{large clamp}, and \textit{foam brick}, while performing worse than the mean on \textit{master chef can}, \textit{tuna fish can}, \textit{bleach cleanser}, and \textit{scissors}.
Interestingly, our methods perform well on identical \textit{large clamp} and \textit{extra large clamp},
whereas both the competing methods perform poorly on these objects.
Real-world robotic applications require handling objects of different sizes and this necessitates highly accurate pose estimates.
The standard procedure of reporting the AUC of ADD-(S) and ADD-S metrics with a fixed threshold of 0.1m does not take the object size into account.
To better reflect the performance of our method on smaller objects, we present the AUC of ADD-(S) and ADD-S metric with a threshold of 10 \% of the object diameter.
We denote this metric as AUC of ADD-(S) and ADD-S @0.1d.
The accuracy of the proposed method drops significantly for smaller objects while using the object-specific threshold.
In particular, the AUC of ADD-(S)@0.1d score for \textit{tuna fish can}, \textit{gelatin box}, and \textit{large marker} are less than 30.
This could be due to the fact that the Pose Loss discussed in~\cref{sec:pose_loss} is computed using the subsampled model points and 
smaller objects contribute less to the overall loss.
In~\cref{tab:inftime}, we also present a comparison of the ADD-S and the mean AUC ADD-S and ADD-(S) scores of the predominant RGB as well as RGB-D methods.
Benefiting from the geometric features imparted by the depth information, RGB-D methods outperform RGB-only methods.
However, RGB-only methods are catching up with the RGB-D methods fast~\citep{sundermeyer2023bop}.

The FFNs in our model generate the set predictions from the decoder output embeddings, which are the result
of cross-attention between the object queries and the encoder output embeddings. 
Each encoder output embedding corresponds to a specific image pixel.
This allows us to investigate the pixels that contribute the most to each object prediction.
In~\cref{fig:multiheadattn}, we visualize the decoder cross-attention corresponding to four different object detections, 
where the attended regions correspond to the object's spatial position in the image very well.
Moreover, looking closely at the pixels with the highest attention score reveals the object parts that contribute most to the object predictions.
For example, in ~\cref{fig:multiheadattn}(a), the tip and base of the~\textit{drill} contribute the most and 
in~\cref{fig:multiheadattn}(d), the spout and the handle~\textit{pitcher base} contribute the most.
Note that in ~\cref{fig:multiheadattn}(a), the base is severely occluded and the base barely visible.
Despite being occluded, the attention mechanism focuses on the base heavily, which demonstrates the significance of the base in~\textit{drill} pose estimation.

In~\cref{tab:improvements}, we present a quantitative comparison of the YOLOPose variant discussed in~\cref{sec:rotest}.
Variant A performs the best among the variants. This can be attributed to the additional 
object-specific information contained in the output embedding.

\begin{table}
      \centering
      \footnotesize
      \setlength{\aboverulesep}{0pt}
      \setlength{\belowrulesep}{0pt}
        \caption{Quantative comparison of the YOLOPose variants. }
\begin{tabular}{l|c|c|c}
    \toprule
    Method & \thead{AUC of \\ ADD(-S)} & \thead{AUC of \\ ADD-S} & \thead{Parameters \\ $\times 10^6$ }\\
    \bottomrule
    YOLOPose  &  82.6 & 90.1 &  \textbf{48.6} \\
    Variant A &  \textbf{83.3} & \textbf{91.2} &53.2 \\
    Variant B  &  82.8 & 91.0 & 53.4 \\
    Variant C  &  82.8 & 90.9 & 52.8 \\
    \bottomrule
  \end{tabular}
    \label{tab:improvements}
\end{table}

\subsection{Inference Time Analysis}
In terms of inference speed, one of the major advantages of our architecture is that the feed-forward prediction networks (FFN) can be executed in parallel
for each object. Thus, irrespective of the number of objects in an image, our model generates pose predictions in parallel. In~\cref{tab:inftime}, we present
the inference time results for 6D pose estimation. 
YOLOPose-A operates at ~45 fps, whereas YOLOpose operates at ~59 fps, which is significantly better than the refinement-based methods
and the RGB-D methods.

\section{Ablation Study}\label{sec:ablation}
In contrast to the standard approach of predicting the 2D keypoints and using a P\textit{n}P solver---which is not trivially differentiable---to estimate the 6D object pose, 
we use the learnable RotEst module to estimate the object orientation from a set of predicted interpolated keypoints.
In this section, we analyze the effectiveness of our RotEst module and the choice of the keypoint representation.

\subsection{Effectiveness of keypoints representations}
We compare various keypoints representations, namely 3D bounding box keypoints (BB), random keypoints sampled using the FPS algorithm, and our representation of choice, the interpolated bounding box keypoints (IBB).
We use the OpenCV implementation of the RANSAC-based EP\textit{n}P algorithm with the same parameters to recover the 6D object pose from the predicted keypoints. Since EP\textit{n}P does not contain any learnable components, this experiment serves the goal of evaluating the ability of the YOLOPose model to estimate different keypoint representations in isolation.
YOLOPose is trained using only the $\ell_1$ loss in the case of BB and FPS representations, 
whereas for the IBB representation, $\ell_1$ is combined with the cross-ratio loss described in~\cref{sec:kploss}.
~\cref{tab:ablation} reports object pose estimation performance for the different representations.
When used in conjecture with the EP\textit{n}P solver, the FPS keypoints performed worse than all other representations. The reason is that the locations of FPS keypoints are less intuitive, making them more difficult to predict, especially for our proposed model that needs to deal with all objects in the YCB-V dataset. In contrast, the IBB keypoints representation yields the best performance. We conjecture that as the cross-ratio loss based on the prior geometric knowledge preserves the keypoints geometrically, this representation is the appropriate choice for our method where a single model is trained for all objects.

\subsection{Effectiveness of RotEst}

\begin{table}
  \centering
  \footnotesize
  \setlength{\aboverulesep}{0pt}
  \setlength{\belowrulesep}{0pt}
  \caption{Ablation study on YCB-V.}
  \begin{tabular}{l|c|c}
    \toprule
     Method & ADD(-S) & \thead{AUC of \\ ADD(-S)} \\
    \midrule
    FPS + EP\textit{n}P & 31.4 & 56.9 \\
    handpicked + EP\textit{n}P  & 31.5 & 55.7 \\
    IBB + EP\textit{n}P & \textbf{56.0} & \textbf{74.7} \\
    \midrule
    IBB + EP\textit{n}P for $R$; head for $t$ & 63.9 & 82.3 \\
    IBB + heads for $R$ and $t$ & \textbf{65.0} & \textbf{82.6} \\
    \bottomrule
  \end{tabular}
  \label{tab:ablation}
\end{table}

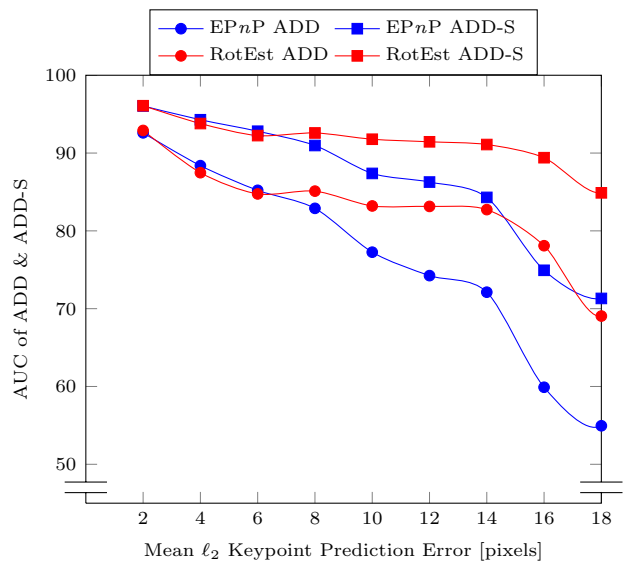
\begin{figure}
    \centering
      \begin{tikzpicture}
\begin{axis}[
    font=\scriptsize,
    xlabel={Mean $\ell_2$ Keypoint Prediction Error [pixels]},
    ylabel={AUC of ADD \& ADD-S},
    xmin=0, xmax=9,
    ymin=45, ymax=100,
    axis y discontinuity=parallel,
    xtick={1,2,3,4,5,6, 7, 8, 9, 10, 11},
    xticklabels={2, 4, 6, 8, 10, 12, 14, 16, 18, 20,  $>$20 },   
    ytick={0,10,...,100},
legend columns=2,
legend style={
          cells={anchor=west},at={(0.5, 1)}, anchor=south,
          font=\scriptsize
      }]

\addplot[smooth,mark=*,blue] plot coordinates {
(1,  92.597752856841 )
(2,  88.382259944920 )
(3,  85.212992040961 )
(4,  82.894157412459 )
(5,  77.260684649802 )
(6,  74.247886752670 )
(7,  72.118704628757 )
(8,  59.901147363463 )
(9,  54.948781398291 )
(10, 67.895930499838 )
(11, 26.329405376394 )
};
\addlegendentry{EP\textit{n}P ADD}

 \addplot[smooth,mark=square*,blue] plot coordinates {
 (1, 96.05430062749397)
 (2, 94.28042502222753)
 (3, 92.80577145916689)
 (4, 90.97090536272936)
 (5, 87.38224299967425)
 (6, 86.27351624107593)
 (7, 84.30140750829307)
 (8, 74.93278086858386)
 (9, 71.31582551561364)
 (10,78.02225004604879)
(11, 43.36474128734435)  
 };
\addlegendentry{EP\textit{n}P ADD-S}

\addplot[smooth,mark=*,red] plot coordinates {
(1, 92.90541896853816  )
(2, 87.47684983938475  )
(3, 84.74889281981458  )
(4, 85.10738768796728  )
(5, 83.20140086513421  )
(6, 83.14158110904692  )
(7, 82.73184669551343  )
(8, 78.09110865333013  )
(9, 69.05265094829188  )
(10,82.43347458516284  )
(11,58.41240889057935  )
};
\addlegendentry{RotEst ADD}

 \addplot[smooth,mark=square*,red] plot coordinates {
 (1, 96.08273006349375)
 (2, 93.79456319420474)
 (3, 92.24393275440146)
 (4, 92.58340200302496)
 (5, 91.77649330563711)
 (6, 91.44701442874547)
 (7, 91.09611788413258)
 (8, 89.40405515932599)
 (9, 84.88760320610849)
 (10,91.62325104626292)
 (11,92.00964206017836)
 };
 \addlegendentry{RotEst ADD-S}

\end{axis}
\end{tikzpicture}
      \caption{Comparison of the pose estimation accuracy with respect to the keypoint estimation accuracy 
      between EP\textit{n}P and RotEst. 
      In the case of highly accurate keypoint estimation, EP\textit{n}P performs comparably to RotEst.
      However, the RotEst module is more robust against inaccuracies in keypoint estimation.
      Overall, RosEst performs better than EP\textit{n}P.
      }
      \label{fig:epnp_rotest_comp}
\end{figure}

After deciding on the keypoint representation, we compare the performance of the learnable feed-forward rotation and translation estimators against 
the analytical EP\textit{n}P algorithm.
The factors that impact rotation and translation components are different~\citep{li2019cdpn}. 
The rotation is highly affected by the object’s appearance in a given image. In contrast, the translation is more vulnerable to the size and 
the location of the object in the image. 
Therefore, we decide to estimate rotation and translation separately. 
In~\cref{tab:ablation}, we report the quantitative comparison of the different variants. 
One can observe that using only the rotation from EP\textit{n}P and directly regressing the translation improved the accuracy significantly. 
In general, RotEst performs slightly better than using EP\textit{n}P orientation and direct translation estimation. 
Furthermore, the RotEst module and the translation estimators are straightforward MLPs and thus do not add much execution time overhead. 
This enables YOLOPose to perform inference in real-time.
Moreover, to quantify the robustness of the RosEst module compared to the EP\textit{n}P algorithm against the inaccuracies in keypoint estimation. 
We exclude the symmetric objects in the comparison. 
\Cref{fig:epnp_rotest_comp}, we present the comparison between the AUC of ADD and ADDS scores achieved by using the RotEst module and using the EP\textit{n}P algorithm for recovering 6D pose from the estimated IBB keypoints. We discretize the average $\ell_2$ pixel error in keypoint point estimation into bins of size two and average the AUC scores for all predictions corresponding to each bin. EP\textit{n}P performs equally well in terms of both the AUC of ADDS metrics compared to the RotEst module when the keypoint estimation accuracy is high.
In the case of large keypoint estimation errors, the RotEst module demonstrates a significantly higher degree of robustness compared to the 
EP\textit{n}P algorithm.

\subsection{Dataset Size-Accuracy Trade-off}
\begin{figure}
    \centering
      \begin{tikzpicture}
   \begin{axis}[
     ybar,
     ymin=0, ymax=1.1,
     bar width=10pt,
     font=\scriptsize,
     tickwidth         = 0pt,
     enlarge x limits  = 0.2,
     xtick={16, 32, 48, 64, 80},
     xtick style={draw=none},
     symbolic x coords = {16, 32, 48, 64, 80},
     nodes near coords,
   legend style={at={(0.25,1.08)},anchor=west},
   legend columns=2,
    ylabel={AUC \&  Cardinality Error },
    xlabel={\# Training Video Sequences},
    width=\linewidth,
    height=.7\linewidth,
    nodes near coords style={anchor=south east,inner sep=0,shift={(4pt, 5pt)},font=\scriptsize,/pgf/number format/.cd,fixed,fixed zerofill},
    ylabel near ticks, ylabel shift={-5pt},
   ]

\addplot coordinates { 
 (16,    .835)
 (32,    .8731)
 (48,    .8732)
 (64,    .8932)
 (80,    .912)};

\addplot coordinates { 
 (16,    .754)
 (32,    .794)
 (48,    .791)
 (64,    .804)
 (80,    .833)};

\addplot coordinates { 
 (16,    .23)
 (32,    .18)
 (48,    .12)
 (64,    .07)
 (80,    .04)};
\legend{AUC of ADD-S, AUC of ADD-(S), Cardinality Error}

\end{axis}
\end{tikzpicture}
      \caption{Comparison of pose estimation and object detection accuracy using a different number of video sequences for training.
      The AUC scores are normalized to the range [0, 1].
      }
      \label{fig:video_subset}
\end{figure}
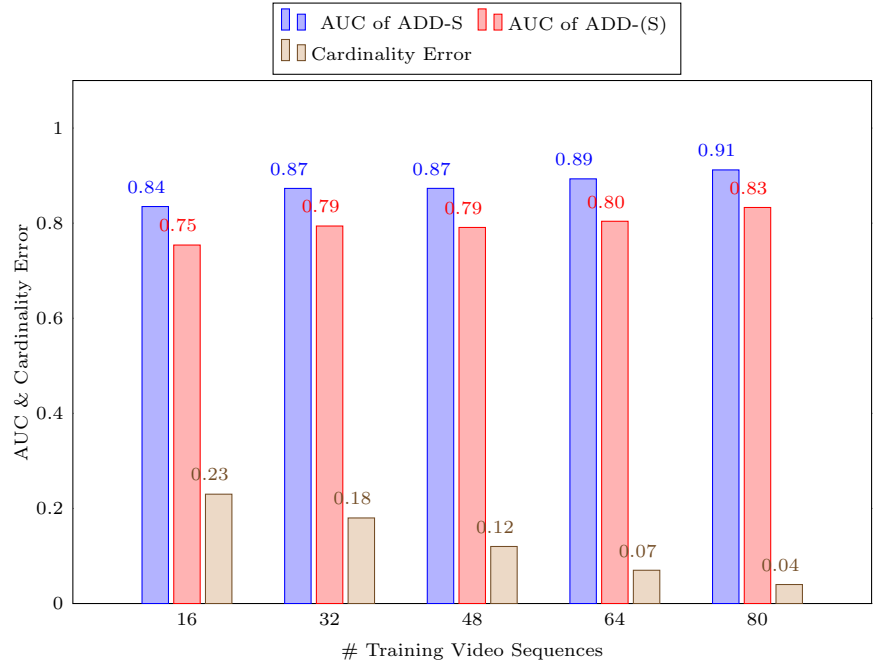

Vision Transformer models match or outperform CNN models in many computer vision tasks, but they require large datasets for pre-training \citep{wang2022towards, cao2022training, Gani_2022_BMVC}.
Furthermore, obtaining large-scale 3D annotations are significantly harder than 2D annotations. Thus, the 3D datasets are supplemented with easy-to-acquire synthetic datasets.
The YOLOPose architecture consists of a CNN backbone model for feature extraction and attention-based encoder-decoder module for set prediction.
Learning set prediction is significantly challenging due to the additional overhead of finding the matching pairs between the ground-truth and the predicted sets
and results in a low convergence rate.
To mitigate this issue, we pre-train our model on the COCO dataset~\citep{lin2014microsoft} for the task of object detection formulated as set prediction.
The COCO dataset comprises 328,000 images with bounding annotations for 80 object categories.
The COCO dataset pre-training enables faster convergence while training on the YCB-V dataset.
To quantify the dataset size-accuracy trade-off in training our model for the task of joint object detection and pose estimation formulated as set prediction, 
we train our model with different subsets of the YCB-V dataset of varying sizes.
As discussed in~\cref{sec:dataset}, YCB-V consists of 92 video sequences. 80 of which are used for training and the rest of them are used for testing.
Additionally,~\citet{xiang2017posecnn} provide 80,000 synthetic images for training as well.
We created five different variants of the training set by using only a subset of the 80 training sequences.
The first variant consists of only 16 video sequences and each subsequent variant consists of 16 additional video sequences added to the previous variant progressively.
All five variants are supplemented with the complete set of synthetic images. 
We train one YOLOPoseA model for each of the dataset variants and evaluate the performance of the models on the test set consisting of twelve video sequences.
In~\cref{fig:video_subset}, we present the AUC of ADD-S and ADD-(S) scores as well as the \textit{cardinality error} (CE), 
which is defined as the $\ell_1$ error between the cardinality of the ground-truth and the predicted set.
The model trained with the smallest training set variant consisting of only 16 video sequences achieves an AUC of ADD-S and ADD-(S) score of 83.5 and 75.4, respectively, whereas
the model trained using the complete training videos achieves an AUC of ADD-S and ADD-(S) score of 91.2 and 83.3, respectively. 
The difference between the models trained using the smallest training set variant and the largest is even more significant in terms of the cardinality error---0.23 compared to 0.04.
This demonstrates the need for large datasets with a wide range of scene configurations.
Presented with smaller datasets with less variability in scene configuration, the YOLOPose model not only performs poorly in terms of pose estimation accuracy but also in terms of object detection accuracy.

\begin{figure}
        \centering
        \newlength{\imgF}
        \setlength{\imgF}{3cm}
        \setlength{\tabcolsep}{0.009cm}
        {
        \begin{tabular}{cccc}
         \includegraphics[width=\imgF]{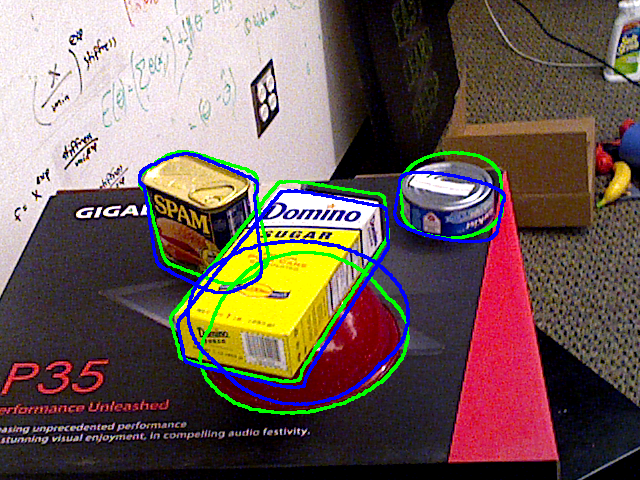} &
         \includegraphics[width=\imgF]{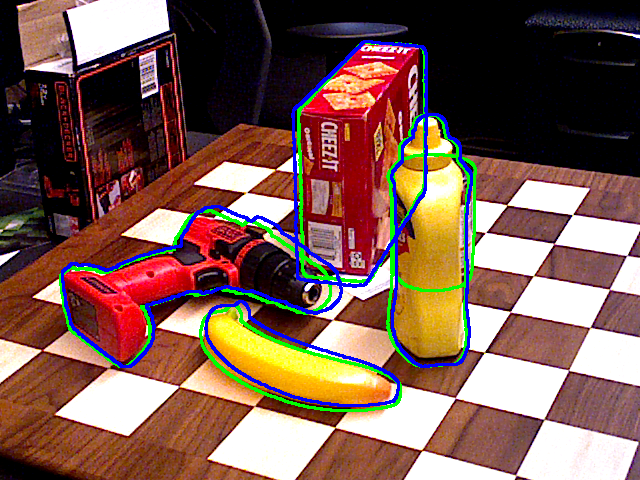} &
         \includegraphics[width=\imgF]{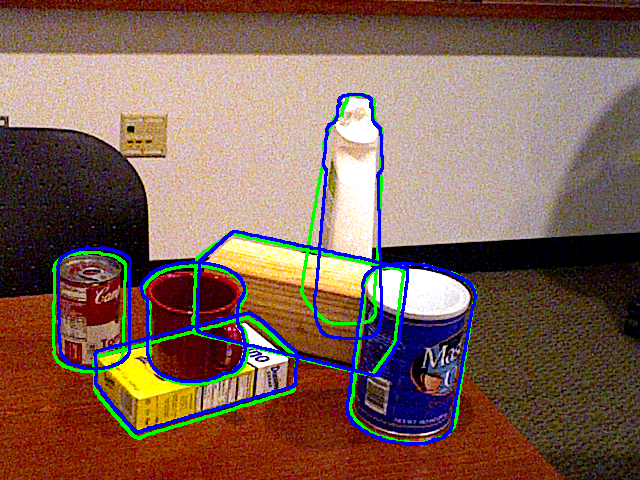} &
         \includegraphics[width=\imgF]{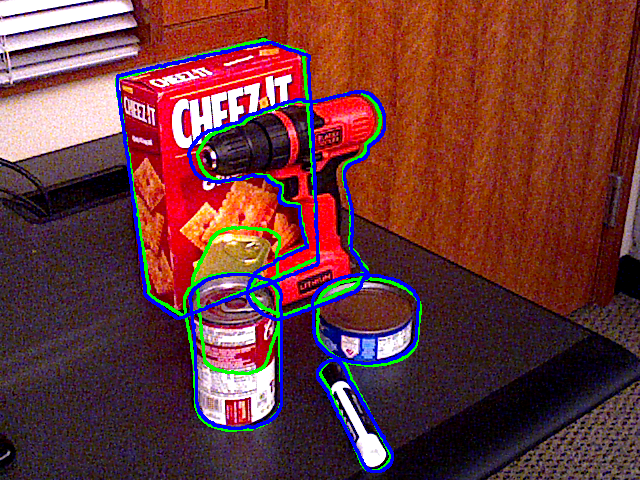} \\
          (a) & (b) & (c) & (d) \\ 
        \end{tabular}
        }
        \caption{Typical failure cases for the YOLOPose model.
        The pose estimation accuracy of our approach is hampered by occlusion.
        Ground truth and predicted object poses are visualized as object contours in green and blue colors, respectively.
        }
        \label{fig:fail_case}
\end{figure}

\begin{figure}
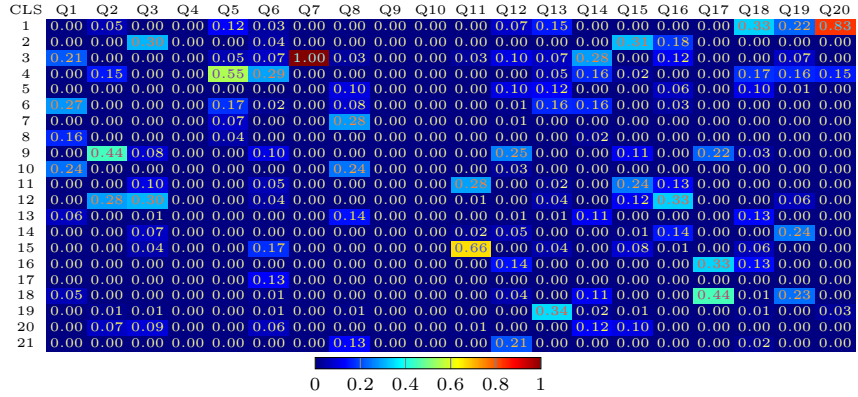

\centering
\pgfplotstabletypeset[
font=\tiny,
    col sep=&,	
	row sep=\\,	
	columns/CLS/.style={reset styles, string type},
 columns/Q1/.style={fixed, fixed zerofill,precision=2},
 columns/Q2/.style={fixed, fixed zerofill,precision=2},
 columns/Q3/.style={fixed, fixed zerofill,precision=2},
 columns/Q4/.style={fixed, fixed zerofill,precision=2},
 columns/Q5/.style={fixed, fixed zerofill,precision=2},
 columns/Q6/.style={fixed, fixed zerofill,precision=2},
 columns/Q7/.style={fixed, fixed zerofill,precision=2},
 columns/Q8/.style={fixed, fixed zerofill,precision=2},
 columns/Q9/.style={fixed, fixed zerofill,precision=2},
 columns/Q10/.style={fixed, fixed zerofill,precision=2},
 columns/Q11/.style={fixed, fixed zerofill,precision=2},
 columns/Q12/.style={fixed, fixed zerofill,precision=2},
 columns/Q13/.style={fixed, fixed zerofill,precision=2},
 columns/Q14/.style={fixed, fixed zerofill,precision=2},
 columns/Q15/.style={fixed, fixed zerofill,precision=2},
 columns/Q16/.style={fixed, fixed zerofill,precision=2},
 columns/Q17/.style={fixed, fixed zerofill,precision=2},
 columns/Q18/.style={fixed, fixed zerofill,precision=2},
 columns/Q19/.style={fixed, fixed zerofill,precision=2},
 columns/Q20/.style={fixed, fixed zerofill,precision=2},
 color cells={min=0,max=1.0,textcolor=-mapped color!80!black},
    /pgfplots/colormap/jet,
]{
CLS&Q1&Q2&Q3&Q4&Q5&Q6&Q7&Q8&Q9&Q10&Q11&Q12&Q13&Q14&Q15&Q16&Q17&Q18&Q19&Q20 \\
1&0.003&0.053&0&0&0.121&0.032&0&0.004&0&0&0&0.072&0.147&0&0&0.002&0&0.331&0.223&0.825\\
2&0&0&0.304&0&0&0.035&0&0&0&0&0&0&0&0&0.311&0.176&0&0&0&0\\
3&0.211&0&0&0&0.051&0.07&1&0.03&0&0&0.029&0.098&0.067&0.28&0&0.115&0&0&0.07&0\\
4&0.003&0.147&0&0&0.545&0.286&0&0&0&0&0&0&0.051&0.164&0.023&0&0&0.171&0.16&0.15\\
5&0&0&0&0&0&0&0&0.095&0&0&0&0.098&0.12&0&0&0.063&0&0.102&0.012&0\\
6&0.272&0&0&0&0.172&0.018&0&0.082&0&0&0&0.01&0.161&0.157&0&0.029&0&0.003&0&0\\
7&0&0&0&0&0.071&0&0&0.277&0&0&0&0.005&0.002&0&0&0&0&0&0&0\\
8&0.161&0&0&0&0.04&0&0&0&0&0&0&0&0&0.023&0&0&0&0&0&0\\
9&0&0.44&0.082&0&0&0.1&0&0&0&0&0&0.253&0&0&0.113&0&0.222&0.031&0&0\\
10&0.235&0&0&0&0&0&0&0.238&0&0&0&0.026&0&0.002&0&0&0&0&0&0\\
11&0&0&0.101&0&0&0.052&0&0&0&0&0.284&0&0.019&0&0.243&0.129&0&0&0&0\\
12&0&0.28&0.304&0&0&0.035&0&0&0&0&0.005&0&0.043&0&0.117&0.329&0&0&0.059&0\\
13&0.063&0&0.005&0&0&0&0&0.139&0&0&0&0.005&0.005&0.113&0&0.002&0&0.13&0&0\\
14&0&0&0.068&0&0&0&0&0&0&0&0.02&0.046&0&0&0.009&0.144&0&0&0.242&0\\
15&0&0&0.043&0&0&0.169&0&0&0&0&0.657&0&0.041&0.002&0.081&0.007&0&0.058&0&0\\
16&0&0&0&0&0&0&0&0&0&0&0&0.139&0&0&0&0&0.333&0.133&0&0\\
17&0&0&0&0&0&0.125&0&0&0&0&0&0&0&0&0&0&0&0&0&0\\
18&0.053&0&0&0&0&0.01&0&0&0&0&0&0.041&0&0.113&0&0&0.444&0.007&0.234&0\\
19&0&0.013&0.005&0&0&0.007&0&0.009&0&0&0&0&0.342&0.021&0.005&0.002&0&0.014&0&0.025\\
20&0&0.067&0.087&0&0&0.06&0&0&0&0&0.005&0&0&0.123&0.099&0&0&0&0&0\\
21&0&0&0&0&0&0&0&0.126&0&0&0&0.206&0&0&0&0&0&0.02&0&0\\
}
\tikz{
            \pgfplotscolorbardrawstandalone[ 
                colormap/jet,
                point meta min=0,
                point meta max=1,
                colorbar horizontal, 
            colorbar style={
                height=0.15cm,
                width=3cm,
                font=\scriptsize
            }]
        }
        \vspace{-5mm}
\caption{Correlation between object queries and the detected object classes. Except for queries 7 and 20, the correlation is weak.
        }
        \label{fig:clsid_q}
\end{figure}

\begin{figure}
\centering
\pgfplotstabletypeset[
font=\tiny,
    col sep=&,	
	row sep=\\,	
	columns/Patch/.style={reset styles, string type},
 columns/Q1/.style={fixed, fixed zerofill,precision=2},
 columns/Q2/.style={fixed, fixed zerofill,precision=2},
 columns/Q3/.style={fixed, fixed zerofill,precision=2},
 columns/Q4/.style={fixed, fixed zerofill,precision=2},
 columns/Q5/.style={fixed, fixed zerofill,precision=2},
 columns/Q6/.style={fixed, fixed zerofill,precision=2},
 columns/Q7/.style={fixed, fixed zerofill,precision=2},
 columns/Q8/.style={fixed, fixed zerofill,precision=2},
 columns/Q9/.style={fixed, fixed zerofill,precision=2},
 columns/Q10/.style={fixed, fixed zerofill,precision=2},
 columns/Q11/.style={fixed, fixed zerofill,precision=2},
 columns/Q12/.style={fixed, fixed zerofill,precision=2},
 columns/Q13/.style={fixed, fixed zerofill,precision=2},
 columns/Q14/.style={fixed, fixed zerofill,precision=2},
 columns/Q15/.style={fixed, fixed zerofill,precision=2},
 columns/Q16/.style={fixed, fixed zerofill,precision=2},
 columns/Q17/.style={fixed, fixed zerofill,precision=2},
 columns/Q18/.style={fixed, fixed zerofill,precision=2},
 columns/Q19/.style={fixed, fixed zerofill,precision=2},
 columns/Q20/.style={fixed, fixed zerofill,precision=2},
 color cells={min=0,max=1.0,textcolor=-mapped color!80!black},
    /pgfplots/colormap/jet,
]{
Patch&Q1&Q2&Q3&Q4&Q5&Q6&Q7&Q8&Q9&Q10&Q11&Q12&Q13&Q14&Q15&Q16&Q17&Q18&Q19&Q20 \\
1&0&0&0&0&0&0&0&0&0&0&0&0&0&0&0&0&0&0&0&0 \\
2&0&0&0.019&0&0&0&0&0&0&0&0&0&0&0&0.023&0&0&0&0&0\\
3&0&0&0.039&0&0&0&0&0&0&0&0&0&0&0&0&0.007&0&0&0&0\\
4&0&0&0&0&0&0&0&0&0&0&0&0&0&0&0&0&0&0&0&0\\
5&0&0&0&0&0&0.022&0&0&0&0&0&0&0&0&0&0&0&0&0&0\\
6&0&0.44&0.261&0&0&0.149&0&0&0&0&0.176&0&0&0&0.932&0&0.222&0&0.016&0\\
7&0&0.013&0.662&0&0&0&0&0.004&0&0&0.127&0&0.027&0&0&0.629&0.111&0.29&0&0.15\\
8&0&0&0&0&0&0&0&0&0&0&0.02&0&0.106&0&0&0.149&0&0.003&0&0.05\\
9&0&0.013&0&0&0&0.281&0&0&0&0&0&0&0&0.044&0&0&0&0&0.012&0\\
10&0.121&0.533&0.01&0&0&0.527&1&0&0&0&0.221&0.418&0&0.442&0.045&0&0.556&0.048&0.953&0\\
11&0.214&0&0.01&0&0.172&0&0&0.71&0&0&0.412&0.557&0.371&0&0&0.112&0.111&0.659&0.012&0.45\\
12&0&0&0&0&0&0&0&0.022&0&0&0.029&0&0.446&0&0&0.102&0&0&0&0.35\\
13&0&0&0&0&0&0.003&0&0&0&0&0&0&0&0.069&0&0&0&0&0&0\\
14&0.169&0&0&0&0&0.018&0&0&0&0&0.015&0.015&0&0.444&0&0&0&0&0.008&0\\
15&0.496&0&0&0&0.616&0&0&0.264&0&0&0&0.01&0.014&0&0&0&0&0&0&0\\
16&0&0&0&0&0.212&0&0&0&0&0&0&0&0.036&0&0&0&0&0&0&0\\
}
\tikz{
            \pgfplotscolorbardrawstandalone[ 
                colormap/jet,
                point meta min=0,
                point meta max=1,
                colorbar horizontal, 
            colorbar style={
                height=0.15cm,
                width=3cm,
                font=\scriptsize
            }]
        }
        \vspace{-5mm}
\caption{Correlation between object queries and the image patch in which the object is detected. The images are divided into 4$\times$4 patches.
Compared to the correlation between object queries and the detected object classes shown in \cref{fig:clsid_q},
the correlation between object queries and image patches is stronger.
        }
        \label{fig:patch_q}
\end{figure}

\begin{figure}
\includegraphics[width=0.85\linewidth]{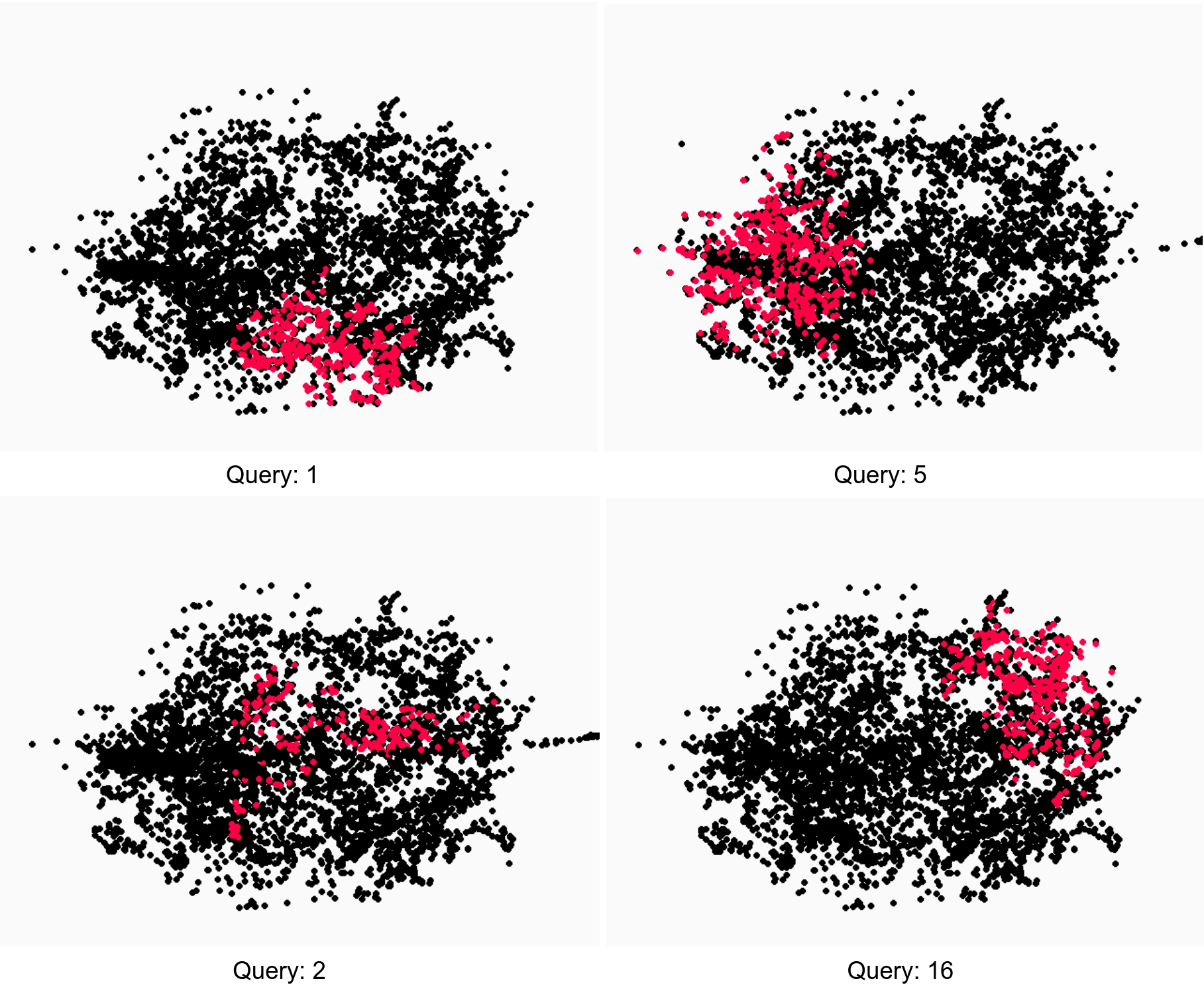}
\caption{Visualization of the center of the bounding boxes predicted by an object query. Black dots represent all the spatial positions
of the ground-truth bounding boxes normalized to the image size present in the test dataset. Red dots represent the bounding boxes predicted by an object query.
Object queries specialize in detecting objects in specific regions of the image.}
\label{fig:q_boxes}
\end{figure}

\section{Understanding Object Queries}
\label{sec:obj_query_us}
To understand the role of the learned object queries in the YOLOPose architecture, we analyzed the correlation 
between the object queries and the detected object class ids as well as the object bounding boxes. 
Since we use only 20 object queries in the YOLOPose architecture---compared to 100 in the DETR~\citep{carion2020end} architecture---we can investigate the object queries individually in detail. 
To this end, we compute the correlation between object queries and class ids, and image patches that form a 4$\times$4 grid.
In \cref{fig:clsid_q}, we visualize the correlation between the object queries and class ids.  Except for queries 7 and 20, the correction is weak.
In contrast, the correlation between the object queries and the image patch of the detected object is stronger (see \cref{fig:patch_q}). 
Note that queries 4, 9, and 10 do not correspond to any objects. This is the case only for the test dataset. 
In the case of the training dataset, all the object queries correspond to object detections.
Moreover, we visualized the spatial location of the center of the bounding boxes predicted by object queries.
In \cref{fig:q_boxes}, we show exemplar visualizations. 
The visualizations also reveal that the object queries specialize in object detection in specific regions of the image.

\section{Limitations}
As shown in~\cref{tab:ycbv-details}, and in~\cref{tab:inftime}, YOLOPose achieves pose estimation accuracy comparable to the state-of-the-art methods.
Despite the impressive accuracy, occlusion remains a big challenge.
In~\cref{fig:fail_case}, we show examples of low-accuracy pose predictions---particularly in the case of partially-occluded objects. 
One of the commonly observed failure cases is the \textit{bowl} object often predicted facing upwards even though \textit{bowl}
is placed downwards (See~\cref{fig:fail_case}a). This is due to the limitation of the symmetry-aware SLoss (\cref{eqn:pose_loss}). 
The SLoss is defined as the $\ell_2$ distance between the closest model points of the object in the predicted and the ground truth poses.
For some objects---\textit{bowl}, for example---the 180° flip error is not penalized enough during training. 

In terms of the dataset needed for training the YOLOPose model, since we formulate the task of joint object detection and 6D pose estimation as a set prediction problem, 
our approach needs pose annotation for all objects in the scene. Some of the commonly used datasets for training and evaluating pose estimation models like
Linemod-Occluded~\citep{Linemodoccluded} and Linemod~\cite{hinterstoisser2013model} that provide pose annotations for just one object per scene in the training set 
cannot be used for training the YOLOPose model.

\section{Discussion \& Conclusion}
We presented YOLOPose, a Transformer-based single-stage multi-object pose estimation method using keypoint
regression. Our model jointly estimates bounding boxes, class labels, translation vectors, and pixel coordinates of 
3D keypoints for all objects in the given input image.
Employing the learnable RotEst module to estimate object orientation from the predicted keypoint coordinates enabled the model to be end-to-end differentiable. 
We reported results comparable to the state-of-the-art approaches on the widely-used YCB-Video dataset and our model is real-time capable. 
Moreover, we presented an improved variant of the YOLOPose model in which the pose estimation FFNs input additional query output embeddings to generate improved pose estimates.
Furthermore, we presented results on the role of object queries in the YOLOPose model. Based on the correlation matrix, we conclude that the object queries
specialize in detecting objects in specific spatial locations.

\section{Acknowledgment}
This work has been funded by the German Ministry of Education and Research (BMBF), grant no. 01IS21080,
project ``Learn2Grasp: Learning Human-like Interactive Grasping based on Visual and Haptic Feedback''.

\bibliography{refs}

\end{document}